\documentclass{applemlr}
\usepackage{amsmath}
\usepackage{enumerate}
\usepackage{algorithm}
\usepackage{algpseudocode}
\usepackage{amsfonts}
\usepackage{amsthm}
\usepackage{cleveref}
\usepackage{diagbox}
\usepackage{colortbl}
\usepackage{amssymb}
\usepackage{xspace}
\usepackage{wrapfig}
\usepackage{adjustbox}
\usepackage{tabularx}
\usepackage{booktabs}
\usepackage{mathtools}
\usepackage{tikz}
\usepackage{enumitem}
\usepackage{silence}
\usepackage{dsfont}
\usepackage[table]{xcolor}
\usepackage[dvipsnames]{xcolor}
\usepackage{multirow}
\usepackage{makecell}
\usepackage{xfakebold}
\usepackage{amsmath,amsfonts,bm}

\def\eqref#1{equation~\ref{#1}}
\def\1{\bm{1}}

\DeclareMathAlphabet{\mathsfit}{\encodingdefault}{\sfdefault}{m}{sl}
\SetMathAlphabet{\mathsfit}{bold}{\encodingdefault}{\sfdefault}{bx}{n}

\definecolor{textgray}{HTML}{6E6E73}
\usetikzlibrary{positioning, calc}
\usetikzlibrary{decorations.pathmorphing}

\makeatletter
\patchcmd{\wrong@fontshape}{\@gobbletwo}{}{}{}
\makeatother
\numberwithin{equation}{section}
\makeatletter
\AtBeginDocument{
  \urlstyle{sf}
  
}
\makeatother

\definecolor{light}{RGB}{125, 125, 125}
\crefname{tcb@cnt@pbox}{code}{code}
\Crefname{tcb@cnt@pbox}{Code}{Code}
\crefname{assumption}{assumption}{assumption}
\Crefname{assumption}{Assumption}{Assumptions}

\newtcolorbox[auto counter]{pbox}[2][]{
  colback=white,
  title=Code~\thetcbcounter: #2,
  #1,fonttitle=\sffamily,
  fontupper=\sffamily,
  arc=2pt,
  colframe=bgcolor,
  coltitle=fgcolor,
  colbacktitle=bgcolor,
  toptitle=0.25cm,
  bottomtitle=0.125cm
}

\makeatletter
\newcommand\applefootnote[1]{%
  \begingroup
  \renewcommand\thefootnote{}%
  \renewcommand\@makefntext[1]{\noindent##1}%
  \footnote{#1}%
  \addtocounter{footnote}{-1}%
  \endgroup
}
\makeatother

\definecolor{cverbbg}{gray}{0.90}

\usepackage{graphicx}
\usepackage{colortbl}
\usepackage{multirow}
\usepackage{multicol}
\usepackage{bm}
\usepackage{subcaption}
\usepackage{diagbox}
\usepackage{amsmath,mathtools}
\usepackage{amssymb}
\usepackage{animate}
\usepackage{setspace}
\usepackage{booktabs} 
\usepackage{tcolorbox}

\usepackage{tabularx}
\usepackage{xspace}
\usepackage{interval}
\usepackage{siunitx}
\usepackage{epigraph}
\usepackage{inconsolata}
\usepackage{caption}
\usepackage{float}
\usepackage{soul}
\usepackage{pifont}
\usepackage{makecell} 
\usepackage{wrapfig}
\usepackage{natbib}
\usepackage{booktabs, multirow, makecell, xcolor, graphicx}
\newcounter{prompt}

\renewcommand{\thefootnote}{\fnsymbol{footnote}}

\definecolor{cvprblue}{rgb}{0.21,0.49,0.74}
\crefname{prompt}{Prompt.}{Prompts.}

\definecolor{lightblue}{rgb}{0.93, 0.96, 1.0}
\definecolor{lightgray}{rgb}{0.92, 0.92, 0.92}
\definecolor{lightgreen}{rgb}{0.89, 0.9375, 0.90625}

\newcommand{\system}{\textit{MintAct}\xspace}

\newcommand{\eg}{e.g.}
\title{\system: A Unified Visual Agent for Digital Environments}

\author{Mingfei Gao$^\circ$$^\dagger$, Rui Tian$^\circ$, Haiming Gang$^\circ$, Bohan Zhai$^\circ$, Le Zhang$^*$, Yuanzheng Gong$^*$, Di Feng$^*$, Ege Özsoy$^*$, \\ Kaixin Ma, Vishwesh Kirthivasan, Oğuzhan Fatih Kar, Roman Bachmann, Anders Boesen Lindbo Larsen, \\ Afshin Dehghan \\

$^\circ$First authors; \,$^*$Core authors; \, $^\dagger$Project lead
}

\vspace{-2pt}
\affiliation{Apple}

\abstract{
We present \system, a family of vision-language models that unifies UI grounding, multi-step navigation across mobile, desktop, and web, and visual tool use,
trained at 2B, 4B, and 8B scales. Through careful design of our environments, data, and training recipes, \system models match the performance of per-domain specialists across all of these capabilities. To enable this, we develop a scalable environment and reinforcement learning (RL) infrastructure. On the environment side, we host hundreds of concurrent instances across heterogeneous per-domain backends, serving both trajectory
data collection and online RL. To enable efficient and scalable RL training, an asynchronous framework keeps explicit control over the cross-domain training distribution and remains stable under noisy environment feedback and off-policy drift. Experimental results show that \system achieves state-of-the-art performance (48.9 on OSWorld-Verified) across a wide range of benchmarks at comparable model sizes.
}  
\date{\sffamily\today}

\begin{document}

%\twocolumn[{
\maketitle
 
\section{Introduction}
\label{sec:intro}

Vision--language agents that operate digital devices directly from pixels are a
promising path toward general digital assistants~\citep{openai2026astra,muse26agent,openai2025cua,google2024gemini2}. To be useful on a real device, an agent must ground instructions to on-screen targets, navigate multi-step
tasks across mobile, desktop, and web interfaces, and reach beyond the screen by
invoking external tools. However, this capability is still fragmented across
specialized models: UI
grounding~\citep{xie2025scalingcomputerusegroundinguser,feizi2025groundingcomputeruseagents},
mobile navigation~\citep{zhang2025appagent,wang2024mobilev2}, desktop
control~\citep{uitars15,wang2025opencuaopenfoundationscomputeruse}, web
navigation~\citep{kar2026weblica,gupta2026molmoweb}, and visual tool
use~\citep{yang2025ultracua,zheng2025deepeyes} are each built, trained, and
benchmarked in isolation. Maintaining a separate specialist per domain is
costly to serve and scale, and especially impractical at the compact sizes
suited for on-device use. This raises a central question: can a single model
unify all of these capabilities without sacrificing per-domain performance?

Unifying them is difficult for reasons that go beyond modeling. The domains
differ in their observation and action spaces, native interactions, data
sources, and executable environments. Naively merging the per-domain action sets
or mixing their data lets domains interfere and erodes per-domain quality.
Moreover, learning reliable interactive behavior increasingly relies on online reinforcement
learning (RL) against heterogeneous environment backends~\citep{zhou2025mai,bai2026asyncwebrl}
that are slow, unreliable,
and produce long multimodal trajectories. Such training is unstable and
resource-hungry, and when domains are mixed under asynchronous execution the
realized training distribution drifts toward whichever domain happens to be
fastest. Therefore, unification is not just a modeling problem, but also a data, environment, and training one.

We present \system, a family of unified and light-weight visual agents at 2B, 4B, and 8B scales. We train \system through a multi-stage recipe that warms up the model with grounding and multi-step supervised fine-tuning, strengthen it with per-domain reinforcement-learning specialists that are distilled back into a
single model, and finally optimize it jointly with agentic RL. Each stage targets a distinct capability gap, and our ablations (\Cref{sec:experiments}) show that every stage contributes. At the center of this recipe is an asynchronous RL framework~\citep{fu2026areal,rastogi2025magistral} that keeps explicit control over the cross-domain training distribution and remains stable under noisy environment feedback and off-policy drift, running on scalable environment infrastructure that serves both online RL rollouts and offline trajectory collection.

Across all three scales, \system models match or exceed per-domain specialists of similar size across grounding, navigation on mobile, desktop, and web,
and visual tool use simultaneously, and are competitive with the best public models at comparable sizes. In particular, \system-8B reaches 48.9 on OSWorld-Verified, 39.1 on Online-Mind2Web, and 67.0 on AndroidWorld. These results show that unification does not need to come at the cost of per-domain quality.

\paragraph{Our contributions are summarized as follows:}
\begin{itemize}
    \item We train \system models at 2B, 4B,
    and 8B scales and unify three capabilities: UI grounding, multi-step
    navigation across mobile, desktop, and web, as well as visual tool use.

    \item We show that jointly training
    across these capabilities and domains does not degrade per-domain performance: a single \system model matches or exceeds size-matched specialist baselines (each trained on an individual domain) across grounding, mobile, desktop, and web navigation, and visual tool use. 

    \item We build an
    asynchronous RL framework tailored to unified visual agents,
    addressing the unstable, heterogeneous environment interaction and the long multimodal trajectories inherent to this setting. This framework
    runs on a scalable environment infrastructure that sustains hundreds of
    concurrent instances across domains. The same environment infrastructure serves both online RL rollouts and offline SFT trajectory collection at high throughput.
    
\end{itemize}

\begin{figure}[t]
    \centering
    % \vspace{-5pt}
    \includegraphics[width=0.9\linewidth]{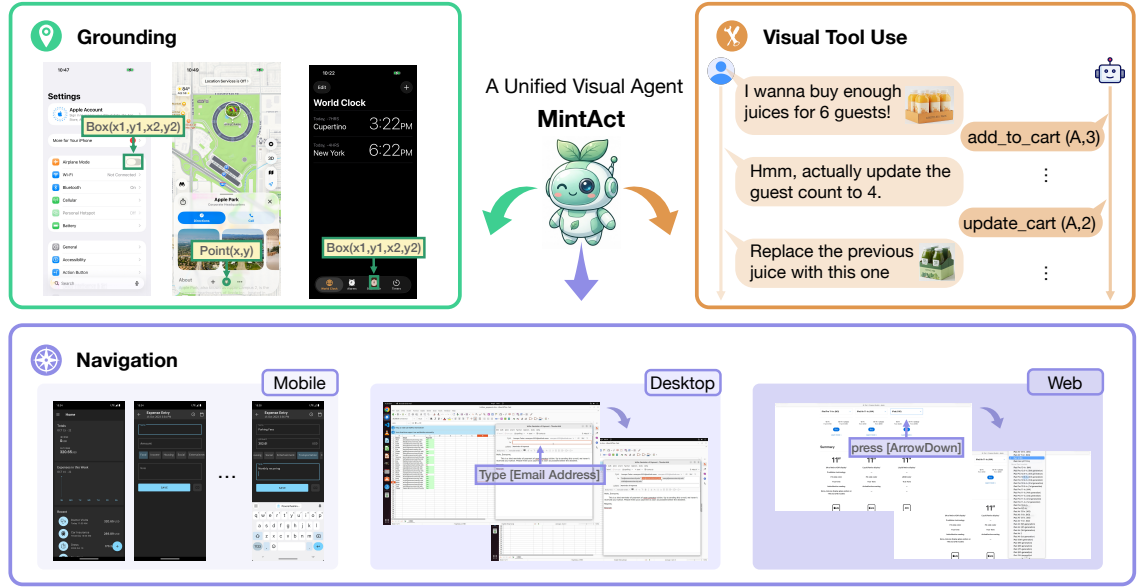}
    \vspace{-5pt}
    \caption{Overview of the diverse visual agentic tasks supported by \system. \system unifies three capabilities including UI grounding, multi-step navigation (across mobile, desktop and web) and visual tool use.}
    \vspace{-5pt}
    \label{fig:mint_va_tasks}
\end{figure}
\section{Related Work}
\label{sec:related}

\subsection{Visual Agents}

With the rapid progress of agentic models, there is growing interest in enabling
multimodal large language models to assist users with everyday workflows, such as
operating mobile applications, browsing the web, and controlling desktop
computers. One promising direction toward such general visual agents is to
interact directly with graphical user interfaces (GUIs), which requires accurate
UI grounding and reliable multi-step navigation. A complementary direction is to invoke external
tools and APIs, grounding information seen in images into structured tool calls
to accomplish tasks that direct UI control cannot easily reach.

\paragraph{GUI grounding and navigation.}
Recent GUI agents improve grounding and navigation through carefully designed action
spaces, instruction tuning with large-scale interaction trajectories, and reinforcement learning. 
For mobile interaction, Ferret-UI Lite~\citep{yang2025ferret} develops a compact on-device agent
through diverse GUI data and reinforcement learning, while MAI-UI targets
real-world deployment with online RL and device-cloud collaboration~\citep{zhou2025mai}.
The Mobile-Agent series progressively incorporates multi-agent planning,
reflection, experience reuse, and scalable environment-based training
to improve complex mobile navigation~\citep{wang2024mobile,wang2024mobilev2}. 
For desktop control, UI-TARS~\citep{uitars15,qin2025ui} and OpenCUA~\citep{wang2025opencuaopenfoundationscomputeruse} advance desktop agents through high-resolution grounding
and long-horizon cross-platform interaction, while EvoCUA~\citep{xue2026evocua} further
integrates task generation, sandbox exploration, and automatic verification
into a self-evolving training loop.
For web navigation, ScaleCUA~\citep{liu2025scalecua} improves cross-site generalization by scaling interaction data from diverse sources, while Weblica~\citep{kar2026weblica} constructs diverse and reproducible web environments through website replay and synthesis, enabling large-scale reinforcement learning for visual web agents. 
Despite this progress, most existing
methods specialize in particular platforms or interaction settings, while
\system unifies grounding and navigation across mobile, desktop, and web
within a single model family.

\paragraph{Visual tool use.}
Beyond direct GUI manipulation, a visual agent can accomplish tasks by grounding
what it sees into calls to structured tools and executable code, reaching
functionality that low-level UI actions cannot easily provide. One line of work
equips models with image-centric operations (cropping, zooming, or running code
over an image) to aid visual reasoning: Visual ChatGPT orchestrates pretrained
visual experts~\citep{wu2023visual}, GPT4Tools learns multi-step tool selection
from tool-use trajectories~\citep{yang2023gpt4tools}, and DeepEyes, OpenThinkIMG,
Thyme, and DeepEyesV2 use reinforcement learning to interleave image operations
with reasoning, code execution, and retrieval~\citep{zheng2025deepeyes,su2025openthinkimg,zhang2025thyme,hong2025deepeyesv2}.
These methods treat tools primarily as aids for \emph{perceiving} a given image.
More recently, hybrid computer-use agents such as MAI-UI and UltraCUA combine
GUI interaction with tools, APIs, and code execution
~\citep{zhou2025mai,yang2025ultracua}, a trend also reflected in benchmarks such as MobileWorld,
OSWorld-MCP, and Agents-last-exam~\citep{kong2026mobileworld,jia2025osworld,sun2026agents}.
Beyond cross-platform GUI
tasks, \system also supports MM-ToolSandBox-style tool use~\citep{ma2026mm}, grounding visual
inputs arriving over multi-turn conversations into high-level actions in
stateful, multi-domain environments.
\subsection{Asynchronous RL for Agents}

Asynchronous RL has emerged as an effective approach for scaling RL post-training
from LLM reasoning to long-horizon interactive agents, where rollout latency is
highly variable and environment interaction is expensive.

\paragraph{Asynchronous RL.} 
RL post-training for LLMs often requires generating many long and highly
variable responses, making synchronous pipelines inefficient as training must
wait for the slowest rollouts in each batch. Asynchronous RL addresses this
bottleneck by allowing rollout workers to generate continuously while trainers
update the policy in parallel. Magistral adopts continuous generation and
frequent in-flight policy synchronization to balance throughput and
on-policyness~\citep{rastogi2025magistral}, while AReaL fully decouples rollout
generation from policy optimization and controls trajectory staleness through
workload balancing and off-policy correction~\citep{fu2026areal}.
AsyncFlow further uses distributed
streaming data management and dynamic load balancing to reduce resource
idling~\citep{han2025asyncflow}.

\paragraph{Multi-turn asynchronous RL.}
As RL is extended from single-turn reasoning to interactive agents, asynchronous
training becomes increasingly important because tool execution and environment
interaction introduce longer and more uneven rollout latency. AgentRL employs a
fully asynchronous pipeline with unified interfaces for multi-turn
tasks~\citep{zhang2025agentrl}, while SkyRL-Agent introduces asynchronous
dispatching for long-horizon tool-using agents~\citep{cao2025skyrl}. Recent
visual-agent systems further adapt this paradigm to costly GUI interaction; for example, AsyncWebRL continuously
overlaps web rollout, policy optimization, and model synchronization through
an everlasting rollout pool~\citep{bai2026asyncwebrl}. These systems, however,
largely target a single domain or modality. \system instead extends asynchronous
RL to the unified, multi-domain visual-agent setting, where heterogeneous
environment backends must be mixed under an explicitly controlled training
distribution while long multimodal trajectories are generated and trained at
scale.

\section{A Unified Visual Agent}
\label{sec:method}

\subsection{Problem Formulation and Overview}
\label{sec:formulation}

\paragraph{Setting.}
Given a natural language instruction $g$ and an observation $o_t$, the agent
follows a policy $\pi_\theta$
\[
  a_t \sim \pi_\theta\big(a_t \mid c,\, g,\, o_{\le t},\, a_{<t}\big),
\] 

which conditioned on a system prompt $c$, the instruction $g$, the history of
observations $o_{\le t}$, and the previous actions with reasoning traces $a_{<t}$, first generates an intermediate reasoning trace and then the next action $a_t$. The action is parsed into
either a UI operation on the screen or a call to an external tool. We
collectively denote this conditioning context, i.e., the system prompt, instruction, observations, and prior reasoning and actions, as the state $s$, and write the policy compactly as $\pi_\theta(a_t \mid s)$.

\paragraph{Capabilities.}
The three capabilities are instances of the general policy above, differing in their horizon and the type of action produced. \emph{Grounding} is a single-step setting in which the agent maps an instruction to a target location on a given screen. The policy reduces to
  \[
    a \sim \pi_\theta\big(a \mid c,\, g,\, o\big),
  \]
where the action $a$ specifies the on-screen coordinates. \emph{Navigation} applies the general multi-step policy above: the agent issues a sequence of interface actions to complete a task across a sequence of screens, on mobile, desktop, and web. \emph{Visual tool use} follows the same multi-step
policy, except that an action $a_t$ may invoke an external tool grounded in the visual context, enabling the agent to solve tasks that require information beyond the screen over multiple steps.

\paragraph{Unified design.}
A single set of weights performs all three capabilities across every domain.
We build a dedicated environment pipeline for each domain including mobile, desktop, web, and visual tool use, together with scalable synthetic environments, and three design choices let one model span them without sacrificing per-domain quality.
\emph{(i)~Shared observation and grounding space.} On every UI domain, the
agent operates purely from raw screenshots and localizes its actions by pixel
coordinates without DOM elements, accessibility trees, or platform-specific
APIs and the coordinates are normalized to a common scale across domains, so
perception and grounding transfer directly across mobile, desktop, and web.
\emph{(ii)~Prompt-conditioned action sets.} To allow a single unified model to handle inherently diverse actions, we avoid naively collapsing the per-domain action sets into a single merged space. Instead, we expose each domain's actions and tools
through a domain-specific system prompt $c$. At inference, the model is
conditioned on the target domain's prompt, which steers it to emit actions from
the corresponding set. \emph{(iii)~Balanced cross-domain mixing.} We keep the
training distribution explicitly balanced across domains throughout both
supervised fine-tuning and reinforcement learning, so that no single domain
dominates and erodes the others. We provide the per-domain action spaces in
\Cref{tab:action_space} and \Cref{sec:env_data}, and the system prompts in
\Cref{sec:appendix_system_prompt}.

\begin{figure}[h]
    \centering
    \includegraphics[width=\linewidth]{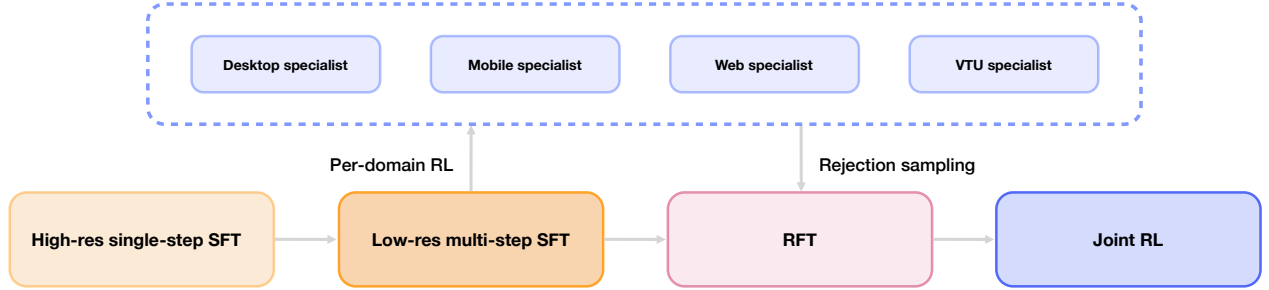}
    \caption{Overview of our training stages. \system starts from a high-resolution single-step SFT for building strong grounding capability followed by a low-resolution multi-step SFT to enable navigation and visual tool use. Then, RL specialists are developed in each domain and distilled back into a single model through RFT. The final stage is joint RL training to further strengthen model performance.}
    \label{fig:training_stages}
\end{figure}

\paragraph{Training overview.}
\system is trained through a sequence of supervised and reinforcement-learning
stages as shown in \Cref{fig:training_stages}. We first elicit basic UI capabilities with high-resolution single-step
SFT, focused on grounding (\Cref{sec:sft_stage1}). We then develop multi-step
interaction with low-resolution multi-step SFT over trajectories generated across
all domains (\Cref{sec:sft_stage2}). Next, we train a per-domain RL specialist
for each domain and distill them back into a single model through an RFT stage with rejection sampling (\Cref{sec:sft_rs}). Finally, we jointly optimize
the policy with agentic asynchronous RL, in which the model interacts directly
with the multiple environments and learns from the outcomes of its own actions
(\Cref{sec:rl}).

\subsection{Environment and Data Pipeline}
\label{sec:env_data}

To strengthen our agent's interaction capabilities, we build executable
environments and scalable data pipelines spanning our domains of interest:
AndroidWorld~\citep{rawles2024androidworlddynamicbenchmarkingenvironment} for
mobile, OSWorld~\citep{OSWorld} for desktop, Weblica~\citep{kar2026weblica} for
web, and MM-ToolSandBox~\citep{ma2026mm} for visual tool use. Complementing these
environments, we introduce scalable synthetic environments spanning operating systems of mobile and desktop that agents can interact with at high
throughput. Agents learn from all of these in two ways: by imitating rollouts
collected from teacher models (SFT), or by interacting with the environments
directly and optimizing against reward signals (RL). Grounding and single-step
capabilities are instead learned from public static datasets, which we describe
with the first SFT stage (\Cref{sec:sft_stage1}).

\paragraph{UI action space overview.}
Each UI domain defines its own action space over raw screenshots, with actions
grounded in pixel coordinates $(x,y)$ normalized to 999$\times$999. \Cref{tab:action_space} summarizes them
side by side: the mobile, desktop, and web action sets cover similar
categories including pointing, scrolling, text entry, navigation, and task
control, but expose domain-specific tokens tailored to each platform's native
interactions. The visual tool-use domain instead acts through structured
function calls and is described separately below.

\begin{table}[t]
\centering
\footnotesize
\setlength{\tabcolsep}{5pt}
\renewcommand{\arraystretch}{1.0}
\caption{Overview of the UI action spaces for the mobile, desktop, and web
domains. Actions are grouped by category. Cells list the concrete action
token(s) defined in each domain, or -- if unavailable. All actions operate on
raw screenshots and are grounded in pixel coordinates $(x,y)$.
$^\dagger$Web has no dedicated \texttt{answer} action. The final response is
returned as part of \texttt{stop}.}
\label{tab:action_space}
\begin{tabular}{@{}l lll@{}}
\toprule
Category & Mobile & Desktop & Web \\
\midrule
\multirow{6}{*}{Pointing}
 & \texttt{click}       & \texttt{left\_click}                           & \texttt{click} \\
 & --                   & \texttt{right\_click}, \texttt{middle\_click}  & -- \\
 & --                   & \texttt{double\_click}, \texttt{triple\_click} & -- \\
 & \texttt{long\_press} & --                                             & -- \\
 & --                   & \texttt{mouse\_move}                           & \texttt{hover} \\
 & --                   & \texttt{left\_click\_drag}                      & -- \\
\midrule
\multirow{2}{*}{Scroll}
 & \texttt{swipe}       & --                                             & -- \\
 & --                   & \texttt{scroll}                                & \texttt{scroll} \\
\midrule
\multirow{2}{*}{Text \& keys}
 & \texttt{type}        & \texttt{type}                                  & \texttt{type} \\
 & --                   & \texttt{key}, \texttt{key\_down}, \texttt{key\_up} & \texttt{press} \\
\midrule
\multirow{2}{*}{Navigation}
 & \texttt{open\_app}   & --                                             & -- \\
 & \texttt{navigate\_back}, \texttt{navigate\_home} & --                 & \texttt{go\_back}, \texttt{go\_forward} \\
\midrule
\multirow{3}{*}{\makecell[l]{Task\\control}}
 & \texttt{wait}        & \texttt{wait}                                  & \texttt{wait} \\
 & \texttt{terminate}   & \texttt{terminate}                             & \texttt{stop}$^\dagger$ \\
 & \texttt{answer}      & --                                             & -- \\
\bottomrule
\end{tabular}
\end{table}

\subsubsection{Desktop}
\label{sec:desktop_env}

\paragraph{Environment.} The environment is a critical component for SFT trajectory generation, RL rollouts, and evaluation. We develop a unified environment pipeline that supports computer-use tasks across all three scenarios. The pipeline decouples the environment from the GPU server and enables communication through HTTP requests. To ensure scalability, we deploy desktop environments on a Linux cluster, with each instance allocated 10 CPU cores and 40 GB of memory. Each environment instance can be hosted independently and configured to serve different purposes. This architecture also improves the fault tolerance of RL training. A failure in a single environment instance does not interrupt the overall training process because the failed instance can be terminated and restarted by the backend while the remaining instances continue generating rollouts. We build our computer environment based on the public OSWorld~\citep{OSWorld}. Each virtual desktop runs inside a Docker container. We connect these containerized environments to our middleware, which exposes HTTP interfaces for communication with external clients. During RL training, the pipeline can concurrently host more than 200 environment instances for online rollouts.

\paragraph{SFT data.}
The data pipeline is designed with two objectives: (i) generating a comprehensive
set of tasks that covers the environment's data distribution, and (ii) collecting
high-quality rollouts for each task for the agent to learn from. Starting from a
small set of human-designed seed tasks, we iterate the following process to
expand coverage. In each round, we prompt a computer-use specialist model (EvoCUA-32B \citep{xue2026evocua}) to roll out the current tasks in the environment using our action space, and then feed the collected rollouts to another strong VLM that proposes new tasks grounded in what the rollouts reveal about the environment. The new tasks seed the next round, and we
repeat until we collected enough tasks and rollouts to cover the domains of the environment. Finally, we apply a VLM-as-judge to evaluate the rollouts and filter out tasks associated with failed rollouts. Thanks to the scalable and efficient design of our environment infrastructure, these rollouts can be collected at high throughput through
parallel execution. EvoCUA natively produces a thinking trace before its final tool calls, which we use directly as our supervision target.

\paragraph{RL data.}
For RL training, we curate tasks from the desktop SFT pool and reuse their
system prompts to best elicit the model's capabilities. To target an appropriate difficulty, we sample eight trajectories per task and retain tasks with a mix of successful and failed rollouts under a VLM judge, favoring those near a 50\% success rate to maximize the learning signal for the group-relative RL objective (\Cref{sec:rl}). This yields roughly 3k OSWorld tasks for online rollouts.

\subsubsection{Mobile}
\label{sec:mobile_env}

\paragraph{Environment.} We build our mobile environment on the open-source AndroidWorld emulator~\citep{rawles2024androidworlddynamicbenchmarkingenvironment}. Each emulator is hosted with the same decoupled, HTTP-based infrastructure as the desktop environment (\Cref{sec:desktop_env}). During RL training, we run more than 100 concurrent instances for online rollouts.

\paragraph{SFT data.}
We follow the same process as in \Cref{sec:desktop_env} to construct meaningful
tasks paired with high-quality rollouts, using a dedicated teacher model to
generate the rollouts and maximize task success rate. However, we find that the
teacher model's thinking traces lack sufficient reasoning detail. To address
this, we add a second, offline thinking-trace labeling pass with a frontier VLM:
given the rollout up to the current step together with the ground-truth action,
the VLM produces the corresponding thinking trace for this step. This relabeling strategy yields
substantially richer reasoning supervision, which we find vital for learning a
capable agent.

\paragraph{RL data.}
Following the same difficulty-based curation as the desktop environment
(\Cref{sec:desktop_env}), we select roughly 3k AndroidWorld tasks with mixed
success and failure under the SFT policy for online RL rollouts.

\subsubsection{Web}
\label{sec:web_env}

\paragraph{Environment.}
 We use the two complementary environment sources introduced by Weblica~\citep{kar2026weblica}: Weblica-Cache and Weblica-Synth. Both are served locally, eliminating network latency, bot-detection failures, and reproducibility issues that plague live-web RL training.
% We reduce action-to-screenshot latency from ~1.5s to 50--150\,ms, enabling fast, parallel rollout generation during asynchronous RL training.

\begin{itemize}
    \item \textbf{Weblica-Cache.}
    Real-world browsing sessions are recorded with Playwright~\citep{playwright}, capturing all HTTP traffic. Rule-based parameter normalization then strips volatile tokens (e.g., dynamic timestamps and session IDs) to ensure deterministic offline replay under complete network isolation.

    \item \textbf{Weblica-Synth.}
    To cover stateful interactions and long-tail domain capabilities, an autonomous coding agent generates self-contained, framework-free HTML, CSS, and JavaScript websites spanning diverse interaction capabilities, domains, and visual designs.

\end{itemize}

\paragraph{SFT data.}
Web navigation queries are sampled across hundreds of thousands of domains from the InstaV3 dataset~\citep{trabucco2025insta}, and Qwen3-VL-32B-Instruct generates execution rollouts on these tasks. Because unconstrained rollouts often contain navigation errors, an offline filtering pass with a strong VLM judge retains only the fully verified successful trajectories ($\sim$51.7k) for supervised warm-starting, providing strong visual grounding and baseline navigation capabilities.

\paragraph{RL data.}
For reinforcement learning, tasks are constructed using Weblica-Synth: thousands of diverse, fully interactive web environments are synthesized across broad navigation capabilities, visual styles, and web domains, each paired with multiple tasks of varying difficulty. This yields a final RL training suite of 10k tasks.

\subsubsection{Visual Tool Use (VTU)} 
\label{sec:vtu}

Beyond direct UI interactions, a visual agent can also invoke structured tools and executable code to reach application functionality that is inefficient, difficult, or impossible to obtain through low-level UI actions alone. To support learning and evaluation of this capability, we build upon the MM-ToolSandBox environment~\citep{ma2026mm} for data generation and agent training.

\paragraph{Environment.}
MM-ToolSandBox provides a stateful simulator for multi-turn and multi-image interactions between users and agents. Images can be introduced progressively throughout a conversation and remain accessible to the agent in subsequent turns, enabling tasks that require cross-image reasoning, visual working memory, and adaptation to evolving user goals. The environment supports both structured tool-use and code-execution interfaces, and contains more than 500 native tools across $16$ application domains. These tools cover a broad range of capabilities, including information retrieval, application operations, visual inspection and manipulation, and system-level utilities.

We primarily adopt the structured tool-use interface to maintain a unified action format with the other environments in our training framework. At each interaction step, the agent produces a structured function call specifying the selected tool and its corresponding arguments. The environment then executes the action, updates its internal state, and returns the resulting textual or visual observation to the agent.

A key challenge of MM-ToolSandBox is its large and hierarchically organized tool space. Exposing the complete inventory of more than 500 tools within every prompt would introduce substantial context overhead and make tool selection unnecessarily difficult. To address this, MM-ToolSandBox supports dynamic tool management through a dedicated \texttt{search\_tool}, which allows the agent to retrieve task-relevant tools on demand and progressively navigate the tool hierarchy, and additionally provides a \texttt{coding\_tool} for operations that are more naturally expressed through code, such as intermediate computation, data transformation, and programmatic inspection of visual inputs. Together, these interfaces allow the agent to combine explicit API invocation with flexible code-based reasoning while operating over a large and diverse action space.

\paragraph{SFT data.} 

To construct diverse training scenarios, we follow the multi-stage generation pipeline proposed in MM-ToolSandBox. First, we perform image--tool association to identify tools and application workflows that can be meaningfully grounded in each visual input. Second, we generate scenario outlines by sampling multiple design dimensions that characterize realistic user--agent interactions. These dimensions include different information-flow patterns, such as aggregating information from multiple images or searching for an entity identified visually, as well as different image-arrival patterns, where images are either provided together at the beginning of the conversation or introduced progressively across multiple turns. This structured sampling enables the generated scenarios to capture the diversity and evolving context of everyday digital-assistant interactions. Third, we instantiate the entities required by each scenario and insert them into the sandbox as state artifacts. These entities define the initial environment state and provide deterministic references for verifying the effects of the agent's actions. Finally, we generate task-specific evaluation rubrics and apply LLM-based quality filtering to remove scenarios that are ambiguous, infeasible, visually ungrounded, or inconsistent with the underlying environment state. We source 100K images from the COYO \citep{kakaobrain2022coyo-700m} and Aria UI \citep{ariaui} dataset for scenario generation to ensure we can get diverse scenarios covering different visual domains and application domains. 

Given the generated scenarios, we use strong proprietary agent models to produce reference interaction trajectories. Each model interacts directly with the environment, receives intermediate tool observations, and continues execution until it completes the task or reaches the maximum interaction budget. We retain only trajectories that pass both complementary evaluation mechanisms: an LLM judge evaluates the trajectory against the task-specific rubrics, while deterministic entity-state checks verify that the expected changes have been correctly applied to the sandbox. This filtering process removes trajectories with incorrect visual grounding, invalid tool calls, incomplete execution, or superficially plausible responses that do not produce the required environment state. 

\paragraph{RL data.}
The scenarios used for reinforcement learning are selected from the broader SFT scenario pool. We run the SFT-initialized policy multiple times on each candidate scenario and prioritize those with an intermediate success rate (centered around $0.5$), for which the current model succeeds on some rollouts but fails on others. Such partially solved scenarios provide the most informative signal for the group-relative RL objective (\Cref{sec:rl}). We additionally balance the selected scenarios across application domains to avoid overrepresenting those that are easier for the initial policy. In total, this yields roughly 800 scenarios for RL training.

\subsection{Supervised Fine-Tuning}
\label{sec:sft}
Starting from a base model, we warm up \system with two
supervised fine-tuning (SFT) stages. A third RFT stage
(\Cref{sec:sft_rs}) further refines the model (see \Cref{fig:training_stages}). All three optimize the standard
autoregressive objective: they maximize $\log \pi_\theta(a \mid s)$, the
log-likelihood of the target reasoning trace and action under the policy
of \Cref{sec:formulation}, where the state $s$ collects the policy's conditioning
context (the system prompt, instruction, and history).

The two warm-up stages
differ in input resolution, task horizon, and data mixture. The first stage
draws on public, static grounding and single-step datasets, whereas the second
stage trains on the multi-step trajectories generated from our per-domain
environments (\Cref{sec:env_data}).

\subsubsection{High-Resolution Single-step SFT}
\label{sec:sft_stage1}
The first stage targets grounding and single-step navigation, both of which
benefit from high input resolution for precise localization. We fine-tune on
grounding data (GroundCUA~\citep{feizi2025groundingcomputeruseagents} and
MolmoPoint~\citep{clark2026molmopoint}) together with single-step navigation
data from GUI-Odyssey~\citep{lu2025guiodyssey} and
AndroidControl~\citep{li2024effects}. For grounding, the state $s$ consists of
a grounding system prompt $c_{\mathrm{g}}$ (a domain-specific instance of $c$), the referring instruction $g$, and
the current screenshot $o$,
\begin{equation}
    s_{\mathrm{ground}} = \big(c_{\mathrm{g}},\; g,\; o\big),
\end{equation}
and the model is trained to generate intermediate reasoning, in
\texttt{\textless grounding\_think\textgreater}\,$\dots$\,\texttt{\textless/grounding\_think\textgreater},
followed by the target coordinates in
\texttt{\textless answer\textgreater}\,$\dots$\,\texttt{\textless/answer\textgreater}.
For single-step navigation, the input additionally includes a textual summary $h$ of the past action history,
  \begin{equation}
    s_{\mathrm{nav}} = \big(c_{\mathrm{n}},\; g,\; h,\; o\big),
  \end{equation}
where $c_{\mathrm{n}}$ is the navigation system prompt. Given this input, the
model predicts the next action in the format
\texttt{Thought:~\dots~Action:~\dots}.
This stage instills the basic perception and grounding capabilities on which the
subsequent stages build. We cap the input at $2{,}116{,}800$ pixels per image and
train for one epoch with a learning rate of $1\times10^{-5}$.
\subsubsection{Low-Resolution Multi-step SFT}
\label{sec:sft_stage2}
The second stage develops the model's multi-step interaction ability for
navigation and visual tool use. Since each trajectory spans a sequence of
screenshots, we lower the input resolution to keep training efficient over long
histories. We train on a mixture of multi-step trajectories drawn from our environments of \Cref{sec:env_data}, grouped into four
domains ($\sim$42.7k for mobile, $\sim$23.5k for desktop, $\sim$51.7k for web, and $\sim$59.8k for visual
tool use). We keep the total number of samples as a fixed budget and
upsample/downsample each domain to a uniform $25\%{:}25\%{:}25\%{:}25\%$ mixing
ratio to keep the domains balanced. Each example is presented in multi-turn format with the system prompt of its domain, so that the model learns to
follow the action set indicated by the prompt. Concretely, at turn $k$ the input
is the domain system prompt $c$ followed by the interleaved history of previous
turns, each consisting of a screenshot, a thought, and an action (including
\texttt{\textless toolcall\textgreater}), together with the current screenshot,
\begin{equation}
    s^{(k)} = \big(c,\; o_1, a_1,\; \dots,\; o_{k-1}, a_{k-1},\; o_k\big),
\end{equation}
and the model is trained to output the current reasoning trace and action $a_k$.
The output format follows the target domain: mobile and web use a
\texttt{Thought:~\dots~Action:~\dots} format, whereas desktop and visual tool use
enclose the reasoning trace in \texttt{\textless think\textgreater\dots\textless/think\textgreater}
before emitting the action.
Unlike the summarized history used in the single-step stage, we append each turn
to the running sequence so the prefix stays unchanged as the trajectory grows.
This append-only format is friendly to the subsequent agentic RL stage, where a
fixed prefix lets the KV cache from earlier turns be reused across turns instead
of recomputed at every step. To keep the sequence length manageable, we retain
trajectories of at most 30 turns for mobile, desktop, and web, and 100 turns for
tool use. We limit the input to 921,600 pixels for each screenshot and train for one epoch with a learning rate of $1\times10^{-5}$.

\paragraph{Trajectory construction for visual tool use.}
As introduced in \Cref{sec:vtu}, our visual agent uses a dynamic tool registry: during execution, the agent can search for new tools and add their definitions to the active function-calling interface. As a result, the tool list in the system prompt changes over the course of a trajectory. This creates a challenge for standard trajectory-based SFT, which assumes a fixed system prompt. Using a single prompt would either expose tools before they are discovered or omit tools required by later actions. 
To address this issue, we introduce tool-registry-aware trajectory segmentation, splitting each trajectory whenever the active tool set changes. Each segment uses the corresponding updated system prompt, ensuring that every supervised action is conditioned on the tools available at that step. In addition, each segment retains the full interaction history for context. Since adjacent segments contain overlapping turns, we apply segment-specific loss masking so that only newly introduced assistant actions contribute to the loss. Thus, every assistant turn is supervised exactly once while remaining conditioned on the complete history and the correct tool registry.

The resulting model is a strong warm-up policy (We refer it as \system-SFT.) that already performs grounding,
navigation, and visual tool use across all domains, and it serves as the starting point
for the reinforcement-learning and RFT stages that follow
(\Cref{sec:sft_rs},~\Cref{sec:rl}).

\subsubsection{RFT with Rejection Sampling}
\label{sec:sft_rs}
The multi-stage SFT produces a single model that already spans all domains. To
further strengthen its per-domain quality, we add an RFT stage driven by rejection sampling~\citep{lu2026thinking,microsoft2025mai}.
Using our RL pipeline (see \Cref{sec:rl_infra}), we train one RL specialist per
domain by having the warm-up model interact with that domain's environment. We
then add an RFT stage that distills these specialists back into a single
model: each specialist generates trajectories on its domain's SFT tasks, a strong
VLM judge filters them to retain only high-quality (successful) trajectories, and
we fine-tune the SFT model on the resulting rejection-sampling data with the
standard SFT objective. This further improves performance across all domains
while keeping a single set of weights.

\paragraph{Data.} For each domain, the specialist rolls out over that domain's
SFT tasks (all or a subset). For mobile, desktop, and web, we collect a single
pass, whereas for visual tool use, we collect four passes per task to obtain
enough correct trajectories. After VLM-judge filtering, this yields roughly
$\sim$37.8k (mobile), $\sim$21k (desktop), $\sim$7k (web), and $\sim$24k (visual tool use) trajectories.

\paragraph{Training.} Starting from \system-SFT, we fine-tune for
one epoch with a small learning rate of $3\times10^{-6}$, balancing the four
domains at a $25\%{:}25\%{:}25\%{:}25\%$ ratio. We refer this model as \system-RFT.

\subsection{Agentic Reinforcement Learning}
\label{sec:rl}

Inspired by recent advances in reinforcement learning for GUI agents~\citep{zhou2025mai,bai2026asyncwebrl}, we further enhance the capabilities of \system through RL-based post-training.
Training
a unified visual agent requires interacting with diverse backends (\eg, Mobile, Web, Desktop) and processing long multimodal trajectories.
To build a highly scalable infrastructure, our system design must address three primary challenges:
\begin{itemize}

\item Slow Environment Interaction:
GUI rollouts require agents to repeatedly execute actions, wait for interface updates, and capture new screenshots in GUI tasks. These environment operations typically take seconds and are substantially slower than model inference. Their latency also varies significantly across different backends.
As a result, conventional synchronous RL suffers from severe GPU underutilization.

\item Long Multimodal Trajectories:
GUI tasks involve multi-turn interactions where each step incorporates high-resolution visual states.
Accumulating these multimodal observations leads to extremely long context windows and massive memory footprints per trajectory, severely bottlenecking both rollout generation and policy training.

\item Controlled Mixed Training:
In mixed-domain training, maintaining a predefined sampling ratio across domains is essential for stable and balanced learning. However, a synchronous pipeline often wastes GPU resources as all workers must wait for the slowest environments to finish. A naive asynchronous pipeline improves throughput, but lets faster environments dominate the data generation and break the required data controllability.

\end{itemize}

To address these challenges, we develop an asynchronous RL framework as below with carefully designed infrastructure (\Cref{sec:rl_infra}) and training strategies (\Cref{sec:rl_algo}).

\subsubsection{Agentic RL infrastructure}
\label{sec:rl_infra}

\begin{figure}[t]
    \centering
    % \vspace{-5pt}
    \includegraphics[width=\linewidth]{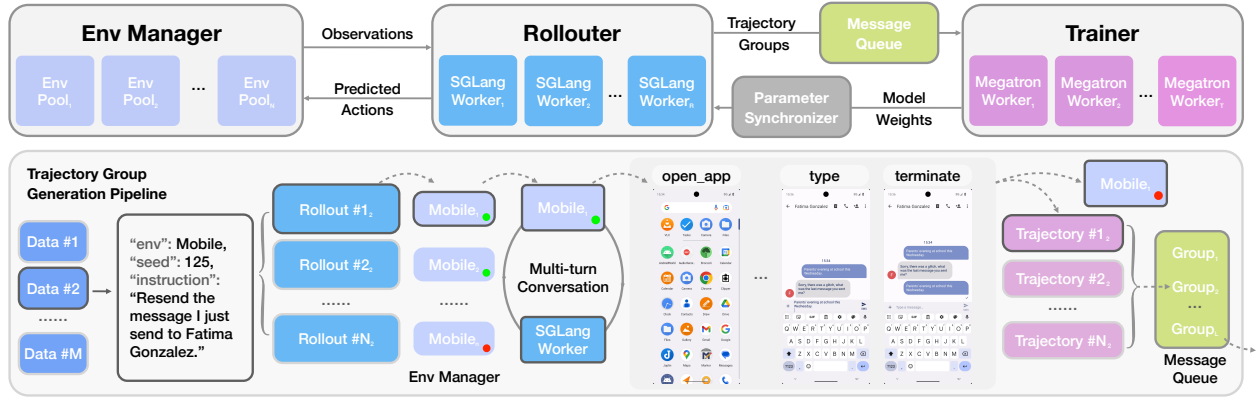}
    \vspace{-5pt}
    \caption{\textbf{Top:} Overview of the proposed asynchronous RL infrastructure. \textbf{Bottom:} Pipeline of trajectory group generation.}
    \vspace{-5pt}
    \label{fig:async_rl_framework}
\end{figure}

Asynchronous RL~\citep{fu2026areal,rastogi2025magistral} decouples rollout generation from policy optimization, allowing both stages to proceed concurrently and reducing idle time caused by slow or uneven trajectory collection.
Building on this paradigm, recent studies have extended asynchronous RL to multi-turn agents~\citep{zhang2025agentrl,cao2025skyrl} and GUI navigation~\citep{bai2026asyncwebrl,li2025efficient}, demonstrating clear gains in resource utilization and training throughput.
As shown in \Cref{fig:async_rl_framework}, we develop an asynchronous RL framework tailored to unified visual agents, built on verl~\citep{sheng2025hybridflow} and rLLM~\citep{rllm2025}. At a high level, the framework forms a closed loop: the \emph{rollouter} generates trajectory groups by interacting with environments provided by the \emph{environment manager}; a \emph{message queue} buffers these groups; the \emph{trainer} consumes them to update the policy; and a \emph{parameter synchronizer} periodically propagates the updated weights back to the rollouter. We describe each component and its design choices below.

\paragraph{Rollouter.}
The Rollouter orchestrates a large number of concurrent rollout processes through an SGLang router~\citep{zheng2024sglang} backed by multiple inference workers, while enforcing a configurable upper bound on the number of active tasks. As the training dataloader iterates over instances from different domains, each instance is replicated $N$ times for group relative policy optimization (GRPO) and assigned to $N$ independent rollout processes, collectively forming a trajectory group.

Each rollout interacts with the corresponding environment backend through the environment manager and performs multi-turn interaction until the agent predicts an end-of-task action (\texttt{terminate}/\texttt{stop}) or reaches the predefined maximum number of turns. At each turn, the latest environment observation is appended to the accumulated interaction context and sent to the SGLang router. The router dispatches each inference request according to the current worker load and KV-cache affinity. The predicted action is then returned to the environment for execution.

\paragraph{Environment manager.}
The environment manager coordinates a large number of heterogeneous environments while providing isolated environment access to concurrent rollout tasks.
It maintains separate environment pools for different backends, including mobile, web, desktop, synthetic, and visual-tool environments. Each rollout acquires an idle environment of the required type from the corresponding pool. The environment is marked as busy and remains exclusively assigned to that rollout throughout the multi-turn interaction, thereby preventing interference from concurrent tasks. Once the rollout terminates, the environment is released and returned to the idle pool for reuse.

\paragraph{Trainer. }
The trainer continuously consumes trajectory groups and their corresponding rewards from the message queue. Once enough groups have been accumulated to form a mini-batch, it computes the advantages and prepares the samples for policy optimization.
In asynchronous training, trajectories may be generated by stale rollout policies, while differences between the rollout and training engines can introduce additional numerical discrepancies even when their model weights are nominally identical.
The trainer therefore reconstructs each trajectory from the token indices and multimodal inputs recorded during rollout. It then computes the proximal-policy log probabilities using the current Megatron~\citep{megatron-lm} actor before optimization.

To accommodate the large memory footprint of long multimodal trajectories, each mini-batch is further partitioned into multiple micro-batches. The trainer performs backward propagation sequentially over these micro-batches and accumulates their gradients before each optimization step.

\paragraph{Message queue. }
The message queue buffers trajectory groups produced by the rollouter and decouples trajectory generation from policy optimization. It follows a first-in, first-out (FIFO) policy, with its maximum capacity defined as a configurable multiple of the training mini-batch size to bound sample staleness. When the queue reaches capacity, the oldest unconsumed trajectory group is evicted upon the arrival of a new group, preventing stale samples from accumulating. The trainer continuously dequeues trajectory groups until enough samples have been collected to form a mini-batch.

Storing the complete multimodal context of each trajectory directly in the queue would impose substantial memory overhead on the host node. We therefore introduce a lightweight multimodal offloading strategy: during rollout, historical images are written to disk, while only their file references and associated metadata are retained in the message queue.
After a trajectory group is dequeued by the trainer, the corresponding images are loaded on demand and materialized as multimodal input tensors on the training devices.

\paragraph{Parameter Synchronizer.}
The Parameter Synchronizer limits the lag between the rollout policy and the latest training policy by periodically propagating the updated trainer weights to the rollout workers.
Specifically, synchronization is triggered every $K$ training steps.
Upon receiving a synchronization signal, the rollouter stops dispatching new rollout tasks, and the SGLang workers finish their in-flight inference requests before weight synchronization begins.
The updated model weights are then transferred from the Megatron training workers to the SGLang inference workers. Once synchronization is complete, inference serving and rollout dispatch are resumed.

Overall, the proposed framework enables scalable and resource-efficient RL training for unified visual agents. It supports hundreds of concurrent interactions across heterogeneous environments, while allowing the number of rollout and training workers to be independently configured to balance their throughputs and minimize GPU idle ratio. Meanwhile, the bounded message queue and periodic parameter synchronization jointly control sample staleness and off-policy magnitude, preserving effective and stable learning signals under asynchronous execution. Together with multimodal offloading and isolated environment management, these designs substantially improve system scalability, resource utilization, and training throughput.

\subsubsection{RL Training Strategies}
\label{sec:rl_algo}
We now describe the strategies used across our RL stages for both the per-domain specialists and the joint model. These apply to any asynchronous RL run: we specify how multi-turn inputs are formed, how informative trajectory groups are selected, and how the policy is optimized stably under noisy environment feedback and off-policy drift.

\paragraph{Multi-turn input format.}
\emph{UI domains.} For UI-domain reinforcement learning, we follow the exact same format as the multi-turn SFT stage, where the concatenation of the system prompt and all preceding screenshots together with predictions serves as the input context for the next agent prediction, preserving the complete multi-turn interaction trajectory throughout the rollout.

\emph{Visual tool use.} We follow the trajectory construction in the  SFT stage including the strategy for dynamic tool registry where the agent can surface additional APIs and thereby rewrite the serialized schemas, at which point we fork the trajectory into separate append-only training segments while the agent continues to condition on the full history. In practice, we only keep the longest fork among all the segments for each trajectory to form the training batch.

\paragraph{Dynamic group filtering.}
We employ a strong VLM as a judge to assign a binary task-completion reward from the task goal and full interaction trajectory.
We adapt DAPO-style over-sampling~\citep{yu2026dapo} to asynchronous training through dynamic filtering of informative groups, removing trajectory groups with all successful or all failed rollouts. The trainer continuously consumes generated groups until a full mini-batch of informative groups is formed. Unlike the offline, difficulty-based task selection in \Cref{sec:env_data}, this filtering operates online, adapting to the evolving success rate and avoiding unnecessary trainer updates on uniform-reward groups.

\paragraph{Off-policy objective and stability.}
Online environments may also encounter system failures, invalid states, or execution timeouts, producing false-negative rewards that do not reflect the agent's actual behavior. We detect trajectories affected by such environment failures and mask them out when computing group statistics and policy losses. Beyond this, two gaps open in a fully asynchronous loop, and they call for different instruments. The
  \emph{staleness} gap separates the current parameters $\theta$ from the parameters
  $\theta^{(v)}$ last pushed to the inference servers. The \emph{engine} gap separates the
  trainer's own forward pass from the serving engine's numerics for the very same tokens. Let
  $m_{i,t}\in\{0,1\}$ be the response mask, which is one only on tokens the policy sampled; let
  $\pi_{\mathrm{prox}}=\pi_{\theta^{(v)}}$ be the trainer's recomputation of the deployed policy and let $\pi_{\mathrm{roll}}$ be the policy whose per-token log-probabilities the inference server returns at
  generation time.

  Advantages come from a group baseline without standard-deviation normalization
  \citep{liu2025understanding}, where $\mathcal{G}(i)$ is the group of $N$ rollouts
  sharing trajectory $i$'s task (\Cref{sec:rl_infra}):
  \begin{equation}
  \hat A_i \;=\; R_i \;-\; \frac{1}{|\mathcal{G}(i)|}\sum_{j\in\mathcal{G}(i)} R_j .
  \label{eq:adv}
  \end{equation}
  We close the staleness gap with an asymmetrically clipped, dual-clipped surrogate. Writing
  $r_{i,t}(\theta)=\exp\!\big(\mathrm{clip}(\log\pi_\theta(a_{i,t}\mid s_{i,t})-\log\pi_{\mathrm{prox}}(a_{i,t}\mid
  s_{i,t}),\,\pm\kappa)\big)$
  and
  $\ell^{\mathrm{hi}}_{i,t}=\max\!\big(-\hat A_i r_{i,t},\,-\hat
  A_i\,\mathrm{clip}(r_{i,t},1-\varepsilon_{\mathrm{lo}},1+\varepsilon_{\mathrm{hi}})\big)$,
  \begin{equation}
  \ell^{\mathrm{clip}}_{i,t} \;=\;
  \begin{cases}
  \ell^{\mathrm{hi}}_{i,t}, & \hat A_i \ge 0,\\[3pt]
  \min\!\big(-c\,\hat A_i,\; \ell^{\mathrm{hi}}_{i,t}\big), & \hat A_i < 0 ,
  \end{cases}
  \label{eq:dualclip}
  \end{equation}
  with $\varepsilon_{\mathrm{lo}}=0.2$, $\varepsilon_{\mathrm{hi}}=0.28$ (clip-higher,
  \citealp{yu2026dapo}), $c=3$, and $\kappa=20$ a numerical guard on the log-ratio. The first branch is
  standard PPO clipping. The second bounds how far a \emph{below-baseline} trajectory can be
  pushed down, which is the direction in which an off-policy update diverges fastest.

  Clipping cannot see the engine gap, because both sides of $r_{i,t}$ are trainer quantities.
  We therefore reweight each token by a truncated importance ratio \citep{yao2025offpolicy} against the distribution that
  actually emitted it,
  \begin{equation}
  \rho_{i,t}=\exp\!\Big(\mathrm{clip}\big(\log\pi_{\mathrm{prox}}(a_{i,t}\mid
  s_{i,t})-\log\pi_{\mathrm{roll}}(a_{i,t}\mid s_{i,t}),\,\pm\kappa\big)\Big),
  \qquad
  w_{i,t}= m_{i,t}\cdot\mathrm{sg}\!\big[\min(\rho_{i,t},\,C)\big],
  \label{eq:tis}
  \end{equation}
  with $C=2$ and $\mathrm{sg}[\cdot]$ a stop-gradient: importance weights change the measure, not
  the objective. The truncation is deliberately one-sided. An upper bound is what stops a handful
  of tokens on which the two implementations disagree from dominating the gradient; no lower bound
  is applied, so a token the trainer considers less likely than the engine did is silently
  down-weighted rather than removed. The objective multiplies the two instruments per token and
  aggregates with a mask-independent normalizer,
  \begin{equation}
  \mathcal{L}(\theta) \;=\; \frac{1}{S}\sum_{i}\sum_{t} w_{i,t}\, \ell^{\mathrm{clip}}_{i,t},
  \qquad S=\min\big(\max_i |y_i|,\;L_{\mathrm{cap}}\big),
  \label{eq:obj}
  \end{equation}
  up to a gradient-accumulation constant introduced by micro-batching, where $L_{\mathrm{cap}}$
  is a fixed cap on the response length. Because $S$ is independent
  of $m$, masking a token removes its contribution rather than redistributing weight onto the
  surviving tokens. We use neither a KL anchor to a reference policy nor an entropy bonus.

\subsubsection{Joint RL Across Heterogeneous Domains}
  \label{sec:jointrl}

  We train \system jointly on tasks drawn from two heterogeneous interaction
  domains: mobile and desktop.
  Training jointly and asynchronously introduces a distributional failure that a single-domain
  pipeline never sees: domains differ by an order of magnitude in rollout latency and in how often
  a rollout group carries any learning signal, so the actual training mixture drifts away. For the optimization side, off-policy failures are shared with single-domain RL
  and are handled by the objective of \Cref{sec:rl_algo}. We control the mixture with the two
  composition mechanisms in \Cref{tab:mech}, which together with the off-policy instruments of
  \Cref{sec:rl_algo} form the four mechanisms of joint asynchronous RL training. 

  \begin{table}[t]
  \centering
  \setlength{\tabcolsep}{4pt}
  \renewcommand{\arraystretch}{1.18}
  \small
  \caption{\textbf{The four mechanisms for improving joint asynchronous RL training.} The top two balance data from different domains. The bottom two control staleness and train/inference mismatch.}
  \begin{tabular}{@{}llll@{}}
  \toprule
  Mechanism & Monitoring Signal & Action & Prevented Failure \\
  \midrule
  Per-domain quota      & domain of each group        & admit or discard the group      & mixture drift \\
  Quota-normalized backpressure & cumulative admitted groups & pause a domain's producers & wasted generation \\
  Dual-clip surrogate   & $\pi_\theta/\pi_{\mathrm{prox}}$ & clip, then floor at $-c\hat A$ & staleness divergence \\
  Truncated IS weight   & $\pi_{\mathrm{prox}}/\pi_{\mathrm{roll}}$ & down-weight the token & engine-mismatch blow-up
  \\
  \bottomrule
  \end{tabular}
  \label{tab:mech}
  \end{table}

\paragraph{Controlling the training composition.}
  A naive mixture is dominated by whichever domain happens to be fastest. The effect compounds
  with group-level filtering: we discard GRPO groups whose rollouts are uniformly correct or
  uniformly incorrect, since their advantages vanish, so a domain's contribution depends not on
  its throughput alone but on its throughput \emph{times} its rate of producing mixed-outcome
  groups.

  \emph{Per-domain quota:} we therefore control composition at the point of consumption rather than production. Producers
  tag every trajectory group with its domain and push it into one shared queue. The trainer drains
  that queue under a per-domain integer quota $q_d$ with $\sum_d q_d = B$ and the number of groups per
  update, and admits a group only if its domain is still below quota. In our main runs $B=64$ with $q_d=32$ for two
  domains and a group size of $8$.

  \emph{Quota-normalized backpressure:} admission control alone discards the fast domain's surplus after paying to generate it, so we
  close the loop back to the producers. Each worker polls the shared queue for the
  \emph{cumulative} number of admitted groups per domain, $c_d$, and forms $r_d = c_d/q_d$, the
  number of quota rounds domain $d$ has supplied. A worker pauses while its own domain leads the
  slowest domain by at least the size of one training batch and resumes once the laggard catches up. 

  \emph{Fallback mechanism:} a domain whose groups are all being discarded as zero-advantage
  never reaches its quota, and the trainer blocks indefinitely. Therefore, after a bounded wait, we relax the per-domain caps and the productive domains backfill
  the batch to full size to continue training.

\section{Experiments}
\label{sec:experiments}

\subsection{Setup}
\label{sec:exp-setup}

\paragraph{Benchmarks.}
We evaluate \system across domains. For \emph{grounding}, we
report MMBench-GUI~\citep{wang2025mmbench}, ScreenSpot-v2~\citep{wu2024atlas}, UI-Vision~\citep{nayak2025uivisiondesktopcentricguibenchmark},
and OSWorld-G~\citep{xie2025scalingcomputerusegroundinguser}. We evaluate \emph{mobile navigation} on AndroidControl~\citep{li2024effects}, AndroidWorld~\citep{rawles2024androidworlddynamicbenchmarkingenvironment}, \emph{desktop navigation} on OSWorld-Verified~\citep{OSWorld}, and \emph{web navigation} on Weblica~\citep{kar2026weblica} and Online-Mind2Web~\citep{xue2025an}. For \emph{visual tool use}, we report on MM-ToolSandBox~\citep{ma2026mm}. We use the official metric for each benchmark to report our results. We set max step to 30 and 100 for online \emph{navigation} and \emph{visual tool use} tasks, respectively.

\subsection{Main Results}
\label{sec:exp-main}
Main results are presented in \Cref{tab:full-comparison}. Across all three scales, \system improves substantially over its Qwen3-VL-Instruct initialization on every domain, and a single \system model is competitive with or outperforms size-matched specialist baselines on grounding, mobile, desktop, and web navigation, and visual tool use simultaneously. At 8B, for instance, \system improves over its Qwen3-VL-8B initialization from $47.6$ to $67.0$ on AndroidWorld, $33.9$ to $48.9$ on OSWorld-Verified, $55.5$ to $74.7$ on Weblica, and $3.1$ to $24.5$ on MM-ToolSandBox, with similar gains at 2B and 4B. Compared with size-matched public models, \system-8B attains the best scores on OSWorld-Verified ($48.9$), Weblica ($74.7$), UI-Vision ($56.6$), and OS-World-G ($64.5$), while remaining competitive on the others (\eg, $67.0$ on AndroidWorld).
  \begin{table}[h]
  \centering
  \caption{Full comparison across model scales. Our numbers are marked in \textbf{bold} or \underline{underline} when they are the best or second-best number compared to baselines.}
  \label{tab:full-comparison}
  \resizebox{\textwidth}{!}{%
  \begin{tabular}{l cccc cc c cc c}
  \toprule
  \multirow{2}{*}{Model}
   & \multicolumn{4}{c}{Grounding}
   & \multicolumn{2}{c}{Mobile Navigation}
   & \multicolumn{1}{c}{Desktop Navigation}
   & \multicolumn{2}{c}{Web Navigation}
   & Toolcall \\
  \cmidrule(lr){2-5}\cmidrule(lr){6-7}\cmidrule(lr){8-8}\cmidrule(lr){9-10}\cmidrule(lr){11-11}
   & \makecell{MMBench\\-GUI} & \makecell{ScreenSpot\\-V2} & UI-Vision & \makecell{OS-\\World-G}
   & \makecell{Android\\Control} & \makecell{Android\\World}
   & \makecell{OSWorld\\-Verified}
   & Weblica & Om2W
   & \makecell{MMTool\\Sandbox} \\
  \midrule
   Qwen3-VL-2B~\citep{bai2025qwen3} & 69.5 & 86.8 & 13.7 & 46.1 & 55.5 & 36.4 & 17.0 & 27.7 & 10.8 & 0.0 \\
  Jedi-3B~\citep{xie2025scalingcomputerusegroundinguser}              & -- & 88.6 & 18.7 & 50.9 & -- & --   & 24.0 & -- & -- & -- \\
  ScaleCUA-3B~\citep{liu2025scalecua}      & -- & --   & --   & 55.7 & -- & 23.3 & 12.4 & -- & -- & -- \\
  Ferret-UI-Lite-3B~\citep{yang2025ferret} & -- & 91.6 & -- & 55.3 & 68.9 & 28.0 & 19.8 & -- & -- & -- \\
  MAI-UI-2B~\citep{zhou2025mai}          & -- & --   & --   & --   & -- & 49.1 & --   & -- & -- & -- \\
  \rowcolor{lightblue}

  \rowcolor{lightblue}
  \system-2B-Final & \textbf{79.2} & \textbf{91.8} & \textbf{43.4} & \textbf{58.2} & \underline{61.2} &	\textbf{59.8} & \textbf{36.1} & \textbf{56.3} & \textbf{25.3} & \textbf{7.8}\\
  \midrule
  \midrule
  Qwen3-VL-4B~\citep{bai2025qwen3} & 82.1 & 92.0 & 27.8 & 58.2 & 64.4 & 45.3 & 26.2 & 47.6 & 22.0 & 0.8 \\
  Qwen3.5-4B~\citep{qwen3_5} & -- & -- & -- & -- & -- & 58.6 & 35.6 & -- & -- & -- \\
  \rowcolor{lightblue}

  \rowcolor{lightblue}
  \system-4B-Final & \textbf{83.6}& \textbf{94.0}& \textbf{56.5}& \textbf{65.3}& \textbf{66.1}& \textbf{62.6}& \textbf{44.8}& \textbf{64.9}& \textbf{31.1} & \textbf{18.9}\\
  \midrule
  \midrule
  Qwen3-VL-8B~\citep{bai2025qwen3} & 80.7 & 92.3 & 25.3 & 61.9 & 65.2 & 47.6 & 33.9 & 55.5 & 26.5 & 3.1 \\
  Jedi-7B~\citep{xie2025scalingcomputerusegroundinguser}          & --    & 91.7  & 24.8 & 54.1 & --   & --   & 27.0 & --   & --   & -- \\
  ScaleCUA-7B~\citep{liu2025scalecua}  & --    & --    & --   & --   & --   & --   & 14.3 & --   & --   & -- \\
  OpenCUA-7B~\citep{wang2025opencuaopenfoundationscomputeruse}   & --    & 92.3  & 29.7 & 55.3 & --   & --   & 26.6 & --   & --   & -- \\
  UI-TARS-1.5-7B~\citep{uitars15}      & --    & --    & --   & --   & --   & --   & 27.5 & --   & 31.3 & -- \\
  MAI-UI-8B~\citep{zhou2025mai}      & 88.8  & --    & 42.4 & 64.2 & 69.1 & 70.7 & --   & --   & --   & -- \\
  EvoCUA-8B~\citep{xue2026evocua}      & --    & --    & --   & --   & --   & --   & 46.1 & --   & --   & -- \\
  WEBLICA-8B~\citep{kar2026weblica}    & 83.7 & 94.5 & --   & --   & --   & --   & --   & 70.6 & 39.2 & -- \\
  Qwen3.5-9B~\citep{qwen3_5}           & --    & --    & --   & --   & --   & 57.8 & 41.8 & --   & --   & -- \\
  \rowcolor{lightblue}

  \rowcolor{lightblue}
  \system-8B-Final & 80.8 & \underline{93.7} & \textbf{56.6} & \textbf{64.5} & \textbf{71.0} & \underline{67.0} & \textbf{48.9} & \textbf{74.7} & \underline{39.1} & \textbf{24.5} \\
  \bottomrule
  \end{tabular}%
  }
  \end{table}

\subsection{Ablation}

\paragraph{Joint SFT model maintains performance from SFT-specialist in each domain.}

As shown in \Cref{tab:sft-specialist-ablation}, our per-domain SFT specialists substantially improve performance in their respective domains, and our joint SFT model stays competitive with each specialist: it matches or exceeds them on mobile, desktop, and visual tool use, and stays within a similar range on grounding and web, while covering all domains in a single model.

  \begin{table}[h]
  \centering
  \caption{Ablation: single-domain SFT specialists vs the jointly-trained \system-SFT-8B. Each specialist is trained only on its
  own domain's data. The joint model is competitive with every specialist, matching or
  exceeding them on mobile, desktop, and tool use, and staying within a similar range on
  grounding and web, while covering all domains in a single model. Best numbers are marked in \textbf{bold}.}
  \label{tab:sft-specialist-ablation}
  \resizebox{\textwidth}{!}{%
  \begin{tabular}{l cccc cc c cc c}
  \toprule
  \multirow{2}{*}{Model}
   & \multicolumn{4}{c}{Grounding}
   & \multicolumn{2}{c}{Mobile Navigation}
   & Desktop Navigation
   & \multicolumn{2}{c}{Web Navigation}
   & Toolcall \\
  \cmidrule(lr){2-5}\cmidrule(lr){6-7}\cmidrule(lr){8-8}\cmidrule(lr){9-10}\cmidrule(lr){11-11}
   & \makecell{MMBench\\-GUI} & \makecell{ScreenSpot\\-V2} & UI-Vision & \makecell{OS-\\World-G}
   & \makecell{Android\\Control} & \makecell{Android\\World}
   & \makecell{OSWorld\\-Verified}
   & Weblica & Om2W
   & \makecell{MMTool\\Sandbox} \\
  \midrule
  Qwen3-VL-8B          & 80.7 & 92.3 & 25.3 & 61.9 & 65.2 & 47.6 & 33.9 & 55.5 & 26.5 & 3.1  \\
  \midrule
  Grounding-specialist & \textbf{84.1} & \textbf{94.6} & \textbf{59.3} & 64.2 & --   & --   & --   & --   & --   & --   \\
  Mobile-specialist    & --   & --   & --   & --   & 69.8 & 62.3 & --   & --   & --   & --   \\
  Desktop-specialist   & --   & --   & --   & --   & --   & --   & 37.7 & --   & --   & --   \\
  Web-specialist       & --   & --   & --   & --   & --   & --   & --   & \textbf{57.9} & 27.0 & --   \\
  Toolcall-specialist  & --   & --   & --   & --   & --   & --   & --   & --   & --   & \textbf{19.8} \\
  \midrule
  \system-SFT-8B       & 82.7 & 94.2 & 56.9 & \textbf{64.7} & \textbf{71.0} & \textbf{63.8} & \textbf{42.6} & 57.5 & \textbf{30.0} & \textbf{19.8} \\
  \bottomrule
  \end{tabular}%
  }
  \end{table}

\paragraph{Multi-stage SFT is essential to keep good performance.} Results are presented in \Cref{tab:multistage-sft-ablation}. Stage~1 is valuable for grounding: on its own it reaches $84.1$ on MMBench-GUI and $59.3$ on UI-Vision, but remains weak on multi-step interaction (e.g., $9.4$ on OSWorld-Verified and $15.6$ on Weblica). Stage~2 is essential for navigation and tool use, recovering OSWorld-Verified to $41.5$ and Weblica to $59.1$, but on its own it degrades grounding (UI-Vision drops to $25.0$). Combining both stages retains strong grounding (UI-Vision $56.9$) while achieving high navigation and tool-use performance, showing that the two stages are complementary.

  \begin{table}[h]
  \centering
  \caption{Ablation: impact of the two SFT stages on \system-8B. Stage~1 is
  high-resolution single-step SFT (grounding-focused); Stage~2 is low-resolution
  multi-step SFT. The last row is the full \system-SFT-8B.}
  \label{tab:multistage-sft-ablation}
  \resizebox{\textwidth}{!}{%
  \begin{tabular}{cc cccc cc c cc c}
  \toprule
  \multirow{2}{*}{Stage 1} & \multirow{2}{*}{Stage 2}
   & \multicolumn{4}{c}{Grounding}
   & \multicolumn{2}{c}{Mobile Navigation}
   & Desktop Navigation
   & \multicolumn{2}{c}{Web Navigation}
   & Toolcall \\
  \cmidrule(lr){3-6}\cmidrule(lr){7-8}\cmidrule(lr){9-9}\cmidrule(lr){10-11}\cmidrule(lr){12-12}
   & & \makecell{MMBench\\-GUI} & \makecell{ScreenSpot\\-V2} & UI-Vision & \makecell{OS-\\World-G}
   & \makecell{Android\\Control} & \makecell{Android\\World}
   & \makecell{OSWorld\\-Verified}
   & Weblica & Om2W
   & \makecell{MMTool\\Sandbox} \\
  \midrule
  \rowcolor{lightgray}
  $\times$ & $\times$ & 80.7 & 92.3 & 25.3 & 61.9 & 65.2 & 47.6 & 33.9 & 55.5 & 26.5 & 3.1 \\
  \checkmark & $\times$ & 84.1 & 94.6 & 59.3 & 64.2 & 70.0 & 45.7 & 9.4 & 15.6 & 2.0 & 0.0 \\
  $\times$ & \checkmark & 80.1 & 93.1 & 25.0 & 56.6 & 60.7 & 65.5 & 41.5 & 59.1 & 25.5 & 21.7 \\
  \rowcolor{lightblue}
  \checkmark & \checkmark & 82.7 & 94.2 & 56.9 & 64.7 & 71.0 & 63.8 & 42.6 & 57.5 & 30.0 & 19.8 \\
  \bottomrule
  \end{tabular}%
  }
  \end{table}

\paragraph{RFT is effective at maintaining domain-specific performance.} As shown in \Cref{tab:enhanced-sft-ablation}, distilling from the per-domain RL specialists improves \system-SFT-8B across most navigation and tool-use benchmarks while keeping a single model. For the mobile domain, the RFT model even surpasses the mobile RL specialist (AndroidWorld $68.1$ vs.\ $67.8$). The same effect holds across model scales: at both 2B and 4B, RFT lifts navigation and tool use over the corresponding SFT baseline while leaving grounding essentially unchanged. For example, AndroidWorld improves from $44.8$ to $55.7$ (2B) and from $53.5$ to $62.4$ (4B), Weblica from $44.4$ to $55.6$ (2B) and from $57.1$ to $65.8$ (4B), and MM-ToolSandBox from $2.7$ to $4.3$ (2B) and from $14.4$ to $18.1$ (4B).

  \begin{table}[h]
  \centering
  \caption{Ablation: RFT with rejection sampling. Each per-domain RL
  specialist starts from \system-SFT-8B and is trained only on its own domain. \system-RFT
  distills them into a single model, matching each specialist on its own domain
  while covering all domains. \system-Final improves on top of \system-RFT via joint RL.}
  \label{tab:enhanced-sft-ablation}
  \resizebox{\textwidth}{!}{%
  \begin{tabular}{l cccc cc c cc c}
  \toprule
  \multirow{2}{*}{Model}
   & \multicolumn{4}{c}{Grounding}
   & \multicolumn{2}{c}{Mobile Navigation}
   & Desktop Navigation
   & \multicolumn{2}{c}{Web Navigation}
   & Toolcall \\
  \cmidrule(lr){2-5}\cmidrule(lr){6-7}\cmidrule(lr){8-8}\cmidrule(lr){9-10}\cmidrule(lr){11-11}
   & \makecell{MMBench\\-GUI} & \makecell{ScreenSpot\\-V2} & UI-Vision & \makecell{OS-\\World-G}
   & \makecell{Android\\Control} & \makecell{Android\\World}
   & \makecell{OSWorld\\-Verified}
   & Weblica & Om2W
   & \makecell{MMTool\\Sandbox} \\
  \midrule
  \multicolumn{11}{@{}l}{\emph{Teacher models: per-domain RL specialists.}} \\
  \midrule
\iffalse % Full-benchmark RL-specialist rows, kept for reference; the visible table shows only each specialist's own-domain metrics.
  RL-desktop & 82.5 & 94.5 & 55.9 & 66.0 & 70.7 & 64.7 & 49.2 & 59.1 & 27.7 & 23.3 \\
  RL-mobile  & 82.6 & 94.1 & 56.7 & 65.8 & 70.5 & 67.8 & 44.2 & 59.8 & 28.4 & 24.7 \\
  RL-web     & 83.4 & 94.5 & 56.5 & 63.8 & 70.5 & 65.1 & 47.0 & 77.1 & 39.8 & 24.8 \\
  RL-VTU     & 83.1 & 94.0 & 56.7 & 64.9 & 71.2 & 63.8 & 44.1 & 60.4 & 27.8 & 22.5 \\
\fi
  RL-desktop & -- & -- & -- & -- & -- & -- & 49.2 & -- & -- & -- \\
  RL-mobile  & -- & -- & -- & -- & 70.5 & 67.8 & -- & -- & -- & -- \\
  RL-web     & -- & -- & -- & -- & -- & -- & -- & 77.1 & 39.8 & -- \\
  RL-VTU     & -- & -- & -- & -- & -- & -- & -- & -- & -- & 22.5 \\
  \midrule
  \system-SFT-2B & 78.7 & 91.4 & 43.7 & 57.3 & 61.7 & 44.8 & 31.0 & 44.4 & 15.7 & 2.7 \\

  \system-RFT-2B & 78.2 & 91.8 & 43.5 & 58.9 & 61.7 & 55.7 & 31.6 & 55.6 & 19.7 & 4.3 \\
  \rowcolor{lightblue}
  \system-2B-Final & 79.2 & 91.8 & 43.4 & 58.2 & 61.2 & 59.8 & 36.1 & 56.3 & 25.3 & 7.8 \\
  \midrule
  \system-SFT-4B & 83.6 & 93.8 & 56.5 & 63.8 & 66.0 & 53.5 & 38.5 & 57.1 & 27.7 & 14.4 \\

  \system-RFT-4B & 83.9 & 94.0 & 56.4 & 64.7 & 64.6 & 62.4 & 41.0 & 65.8 & 27.9 & 18.1 \\
  \rowcolor{lightblue}
  \system-4B-Final & 83.6 & 94.0 & 56.5 & 65.3 & 66.1 & 62.6 & 44.8 & 64.9 & 31.1 & 18.9 \\
  \midrule
  \system-SFT-8B & 82.7 & 94.2 & 56.9 & 64.7 & 71.0 & 63.8 & 42.6 & 57.5 & 30.0 & 19.8 \\
  \system-RFT-8B & 81.7 & 93.9 & 56.3 & 64.9 & 70.9 & 68.1 & 43.1 & 72.8 & 36.3 & 21.3 \\
  \rowcolor{lightblue}
  \system-8B-Final & 80.8 & 93.7 & 56.6 & 64.5 & 71.0 & 67.0 & 48.9 & 74.7 & 39.1 & 24.5 \\
  \bottomrule
  \end{tabular}%
  }
  \end{table}

\paragraph{Joint RL model outperforms RL-specialists optimized for individual environments on average.}
To isolate the impact of joint RL, we start from the warm-up SFT checkpoint and jointly optimize on the mobile and desktop environments. Results are in \Cref{tab:rl-specialist-ablation}. The jointly trained model improves substantially over the SFT model and, on average, exceeds every single-domain RL specialist on both grounding and navigation. Training on mobile and desktop together also transfers to the web domain (Weblica from $57.5$ to $60.8$, Om2W from $30.0$ to $32.8$), even though no web data is used in this stage. Visual tool use, the most distant domain, is not part of joint training and stays at the SFT level, suggesting that transfer is easier across nearby domains than distant ones.

  \begin{table}[h]
  \centering
  \caption{Ablation: single-domain RL specialists vs \system-RL-8B. Each specialist is RL-trained only on its own domain's
  environments, starting from the same \system-SFT-8B initialization. The joint
  model is RL-trained across mobile and desktop domains. Grounding Avg.\ averages the
  four grounding benchmarks. Nav.\ Avg.\ averages the five navigation benchmarks
  (AndroidControl, AndroidWorld, OSWorld-Verified, Weblica, Om2W).}
  \label{tab:rl-specialist-ablation}
  \resizebox{\textwidth}{!}{%
  \begin{tabular}{l cccc c cc c cc c c}
  \toprule
  \multirow{2}{*}{Model}
   & \multicolumn{4}{c}{Grounding}
   & \multirow{2}{*}{\makecell{\textbf{Ground.}\\ \textbf{Avg.}}}
   & \multicolumn{2}{c}{Mobile Navigation}
   & Desktop Navigation
   & \multicolumn{2}{c}{Web Navigation}
   & \multirow{2}{*}{\makecell{\textbf{Nav.}\\ \textbf{Avg.}}}
   & Toolcall \\
  \cmidrule(lr){2-5}\cmidrule(lr){7-8}\cmidrule(lr){9-9}\cmidrule(lr){10-11}\cmidrule(lr){13-13}
   & \makecell{MMBench\\-GUI} & \makecell{ScreenSpot\\-V2} & UI-Vision & \makecell{OS-\\World-G}
   &
   & \makecell{Android\\Control} & \makecell{Android\\World}
   & \makecell{OSWorld\\-Verified}
   & Weblica & Om2W
   &
   & \makecell{MMTool\\Sandbox} \\
  \midrule
  Qwen3-VL-8B-Instruct~\citep{bai2025qwen3} & 80.7 & 92.3 & 25.3 & 61.9 & 65.1 & 65.2 & 47.6 & 33.9 & 55.5 & 26.5 & 45.7 & 3.1 \\
  \rowcolor{lightblue}
  \system-SFT-8B & 82.7 & 94.2 & 56.9 & 64.7 & 74.6 & 71.0 & 63.8 & 42.6 & 57.5 & 30.0 & 53.0 & 19.8 \\
  \midrule
  RL-mobile   & 82.6 & 94.1 & 56.7 & 65.8 & 74.8 & 70.5 & 67.8 & 44.2 & 59.8 & 28.4 & 54.1 & 24.7 \\
  RL-desktop  & 82.5 & 94.5 & 55.9 & 66.0 & 74.7 & 70.7 & 64.7 & 49.2 & 59.1 & 27.7 & 54.3 & 23.3 \\
  \rowcolor{lightblue}
  RL-joint & 83.5 & 94.6 & 56.6 & 64.9 & \textbf{74.9} & 70.7 & 66.7 & 48.3 & 60.8 & 32.8 & \textbf{55.9} & 19.8 \\
  \bottomrule
  \end{tabular}%
  }
  \end{table}

\paragraph{Joint RL improves performance even after the RFT stage.}
We compare \system-RFT with the final joint-RL model across scales in \Cref{tab:enhanced-sft-ablation}. Although joint RL optimizes only the mobile and desktop domains (\Cref{sec:jointrl}), it improves most online benchmarks while leaving grounding essentially unchanged (all grounding scores move within about one point). At 8B, \emph{OSWorld-Verified} rises from $43.1$ to $48.9$, and the gains transfer to web (\emph{Weblica} from $72.8$ to $74.7$ and \emph{Om2W} from $36.3$ to $39.1$) against a minor $1.1$ point drop on AndroidWorld. The trend holds at 4B (OSWorld-Verified from $41.0$ to $44.8$, Om2W from $27.9$ to $31.1$) and is strongest at 2B (OSWorld-Verified from $31.6$ to $36.1$, Om2W from $19.7$ to $25.3$).

\paragraph{Synthetic environment helps when no enough exposure to real data.} We also explored a direction of building extensive synthetic environments for mobile and desktop, and optimize our policy by interacting with it (See \Cref{sec:synthetic} for details). This is a scalable way to create massive training data stably. \Cref{tab:synthetic-ablation} shows the contribution of our synthetic environments by applying
the same synthetic RL stage to two initializations: the base Qwen3-VL-8B model and
\system-SFT-8B.

\noindent\emph{Synthetic environments significantly improves its in-domain benchmarks.} \emph{Synthetic Mobile} increases from $26.9$ to $43.1$ after \system-SFT and to $51.9$ after synthetic RL, and \emph{Synthetic Desktop} from $39.3$ to $50.0$ after \system-SFT and to $66.4$ after synthetic RL.

\noindent\emph{Synthetic environments can transfer to real benchmarks.} Training the
base model purely in synthetic environments, without any real interaction data, improves
AndroidWorld from $47.6$ to $60.3$ and OSWorld-Verified from $33.9$ to $38.6$. Neither the real
AndroidWorld nor OSWorld emulators are used at any point in this stage, so
these gains come entirely from skills acquired in synthetic environments and carried over to the
real ones.

\noindent\emph{Gains are more significant in domains with less prior exposure.} On the domains where the prior exposure of the SFT model is more limited, performance rises consistently from the base model through SFT to synthetic RL:
On the iOSWorld~\citep{keunho2026iosworld} benchmark, \system-SFT, trained without any real iOS data, reaches $12.8$ compared to the base model's $2.3$.
Synthetic environments alone raise it to $18.8$ and applying synthetic RL on top of \system-SFT-8B compounds further to $22.3$.

\noindent\emph{On the other hand, when in-domain real data is available, synthetic environments lose
their advantage.} As shown by the \emph{+ RL-joint} row, real in-domain RL yields a larger gain on the
in-domain benchmark (AndroidWorld $66.7$ vs.\ $59.8$ and OSWorld $48.3$ vs $39.4$ from synthetic RL alone).

  \begin{table}[h]
  \centering
  \caption{Ablation: impact of synthetic environments on \system-8B. We compare
  RL on synthetic environments from different initializations (Qwen3-VL-8B vs.\ \system-SFT-8B). RL-joint indicates training jointly on real environments (AndroidWorld and OSWorld).}
  \label{tab:synthetic-ablation}
  \resizebox{\textwidth}{!}{%
  \begin{tabular}{l ccc cc cc c}
  \toprule
  \multirow{2}{*}{Model}
   & \multicolumn{3}{c}{Mobile Navigation}
   & \multicolumn{2}{c}{Desktop Navigation}
   & \multicolumn{2}{c}{Web Navigation}
   & Toolcall \\
  \cmidrule(lr){2-4}\cmidrule(lr){5-6}\cmidrule(lr){7-8}\cmidrule(lr){9-9}
   & \makecell{Android\\World} & iOSWorld & \makecell{Synthetic\\Mobile}
   & \makecell{OSWorld\\-Verified} & \makecell{Synthetic\\Desktop}
   & Weblica & Om2W
   & \makecell{MMTool\\Sandbox} \\
  \midrule
  Qwen3-VL-8B-Instruct~\citep{bai2025qwen3} & 47.6 & 2.3 & 26.9 & 33.9 & 39.3 & 55.5 & 26.5 & 3.1 \\
  + RL-synthetic      & 60.3 & 18.8 & 45.6 & 38.6 & 67.9 & 64.0 & 30.2 & 9.3 \\
  \midrule
  \system-SFT-8B & 63.8 & 12.8 & 43.1 & 42.6 & 50.0 & 57.5 & 30.0 & 19.8 \\
  + RL-synthetic & 59.8 & 22.3 & 51.9 & 39.4 & 66.4 & 66.4 & 28.9 & 21.1 \\
  + RL-joint & 66.7  & 24.8 & 43.8 &  48.3&  56.4&  60.8&  32.8&  19.8 \\
  \bottomrule
  \end{tabular}%
  }
  \end{table}

\section{Conclusion}
\label{sec:conclusion}

We presented \system, a family of visual agents at 2B, 4B, and 8B scales that
unify UI grounding, multi-step navigation across mobile, desktop, and web, and
visual tool use within a single model. Through careful design of our
environments, data, and training, \system models match or exceed
size-matched per-domain specialists while covering all of these capabilities at
once, showing that unification does not come at the cost of per-domain quality.

Central to this result are a scalable RL
infrastructure and an effective training recipe. The RL infrastructure hosts hundreds of concurrent instances across
heterogeneous per-domain backends, keeps explicit control over the cross-domain
training distribution, and remains stable under noisy environment feedback and
off-policy drift. The multi-stage training recipe warm-starts the model,
sharpens it with per-domain RL and rejection-sampling distillation, and finally
optimizes it across domains via joint RL. Synthetic environments additionally offer a complementary, low-cost source of training data for domains where real interaction data is scarce. The environments, infrastructure, and training recipe presented in this work provide a foundation for further scaling unified visual agents.

\section{Limitation}
\label{sec:limitation}

While \system unifies grounding, navigation, and visual tool use in a single
model, several directions remain open.

\paragraph{Scaling joint RL to all environments.} Our experiments show that joint
RL across mobile and desktop is beneficial, improving over the single-domain
specialists on average. Extending joint RL to all available environments,
covering mobile, desktop, web, and visual tool use together, is a promising
direction for further gains. However, doing so is resource-intensive, as it
requires sustaining many heterogeneous environments concurrently throughout
training, and we leave it for future work.

\paragraph{Context management for long-horizon interaction.} During training, we
append every incoming screenshot to the context. This is the most
straightforward setup for RL, but it leads to rapidly growing context and large
memory usage as trajectories lengthen. More sophisticated context management and
agent harnessing, for example summarizing or pruning stale observations, are a
potential future direction for scaling to longer-horizon tasks.

\paragraph{Unifying navigation and tool use.} We include visual tool use in
\system, but it is currently treated largely as a separate capability. Ideally,
a single model would automatically switch between pixel-level navigation and
dynamic tool calling whenever appropriate, invoking tools only when they are
more effective than direct interaction. Achieving this seamless, on-demand
switching within one policy is an important direction for future work.

\paragraph{On-device and server-side collaboration.} Our models range from 2B to
8B parameters, a scale well suited to on-device deployment. A promising direction
is to study collaboration between on-device and server-side models, where a
compact on-device model handles simple and privacy-sensitive tasks locally and
hands off to a larger server-side model only when a task exceeds its capacity.
This could combine the latency and privacy benefits of local execution with the
capability of large-scale models.

\paragraph{Function-calling formulation.}
We follow the standard function-calling formulation, where available tool definitions are provided as part of the model context~\citep{gpt5-function-calling-api,gemini-function-calling-api,anthropic-function-calling-api}. With a dynamic tool registry, however, newly discovered tools must be introduced during execution, changing the prompt prefix and reducing the effectiveness of prefix caching. Our current training formulation handles this through trajectory segmentation: each segment is treated as an independent training example with the corresponding tool registry. For RL, we further assign the trajectory-level reward to its segments and use the last segment for training. We find including earlier segments with random sampling increases training stability, but leads to similar results with more compute. While effective in practice, this provides only coarse credit assignment and is an approximation of the original end-to-end interaction.

A more elegant solution would require function-calling interfaces that natively support dynamic tool discovery without reconstructing the preceding context. Recent systems have begun exploring such formulations through deferred or dynamically loaded tool definitions~\citep{kimi-function-calling-api}. Alternatively, coding-agent interfaces can expose a stable execution environment through which tools are discovered and invoked programmatically. We leave these formulations, together with more principled segment-level credit assignment, to future work.

\section{Acknowledgment} We thank Peter Fu, Jun Wang, Angie Wang, Wencong Zhang, Michael Feng, Yena Han, Dongxu Li, Wentao Wu, Aleksei Timofeev, Zhen Yang, Eldon Schoop, Jeff Nichols, Omar Attia, Andrew Szot, Zhe Gan for their help.
% \clearpage 

\small \bibliographystyle{plainnat} 
\bibliography{main}

\clearpage
\setcounter{page}{1}
%\maketitlesupplementary
\renewcommand{\thetable}{\Alph{table}}
\setcounter{table}{0}
\renewcommand{\thefigure}{\Alph{figure}}
\setcounter{figure}{0}
\appendix

\section{Synthetic Environments}
\label{sec:synthetic}
Interactive environments, such as the live web and device emulators like AndroidWorld~\citep{rawles2024androidworlddynamicbenchmarkingenvironment} and
OSWorld~\citep{OSWorld}, are indispensable for evaluation but costly to scale for training: each rollout ties up
a live page, emulator, or virtual machine. We
therefore also explore a different route and build our training environments \emph{synthetically},
through coding agents. Each environment is created using web technologies (HTML, CSS, and JavaScript) and rendered in a headless browser. The resulting environments are lightweight, deterministic and durable, and run
without any emulator or virtual machine, allowing us to host thousands of environments directly on the training cluster without any external dependencies. These environments are also significantly faster to interact with,
allowing us to collect more rollouts per unit time. We build on the synthetic-website generation of
Weblica~\citep{kar2026weblica} and extend it from single web pages to complete operating-system
interfaces, covering both mobile and desktop.

\begin{figure}[h]
    \centering
    \includegraphics[width=\linewidth]{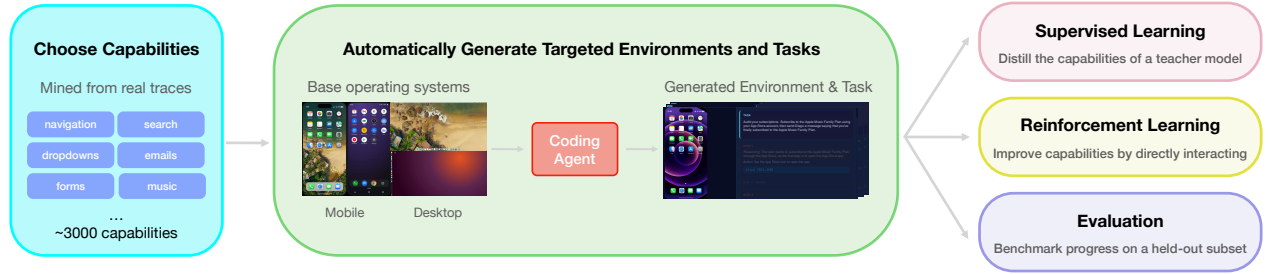}
    \caption{We first choose the capabilities to target, mining
    them from real interaction traces and clustering them into groups. For each capability, a
    coding agent forks one of our four base operating systems and adapts it, together with
    accompanying tasks. The generated
    environments then serve three purposes: distilling a teacher model through supervised
    learning, improving the agent through direct interaction with reinforcement learning, and
    benchmarking progress on a held-out subset of capabilities.
    }
    \label{fig:synthethic_visual}
\end{figure}

\paragraph{Capability flywheel.}
We generate these environments through a \emph{capability flywheel}, a fully automated loop of
three stages. \emph{(i)~Choose capabilities.} We first decide which skills to
target, for example navigation, search, dropdown selection,
composing an email, or playing music. While these capabilities can be specified by hand, in this work we
derive them from data by analyzing the distribution of interactions in large corpora of real
usage traces. \emph{(ii)~Generate environments and tasks.} For each target capability, a coding
agent automatically generates environments and accompanying tasks across our synthetic domains, iterating until the tasks are solvable in the
generated environment. \emph{(iii)~Learn and evaluate.} The resulting environments serve three
purposes: supervised fine-tuning, where we train our agent to imitate ground-truth trajectories
produced by a stronger teacher model; reinforcement learning, where the agent improves by
interacting with the environments directly and optimizing against task rewards; and evaluation,
where we benchmark progress on a held-out subset of
capabilities.

\paragraph{Capability extraction.}
Extraction proceeds in two passes. A frontier model takes as input each trajectory and names the
fine-grained capabilities it covers (e.g., using a modal sheet or scrolling a list),
over-generating candidates to favor recall. A second pass then merges synonyms and clusters them
into higher-level groups, such as form filling, cart management and checkout, setting reminders,
or toggling a setting. The trajectories come from existing interaction corpora:
Android-in-the-Wild~\citep{rawles2023androidinthewild} and MONDAY~\citep{jang2025scalable} for the mobile platforms and
AgentNet~\citep{wang2025opencuaopenfoundationscomputeruse} for the desktop platforms. 

\paragraph{Environment and task generation.}
Each platform starts from a \emph{base} environment: a full operating-system shell with its core
apps that coding agents generate over several human-in-the-loop iterations and that we vet for
fidelity and quality. To cover a capability group, we fork this base and adapt it so the
capability can be exercised (e.g., adding a checkout flow to a store app), together with a small
set of tasks of graded difficulty. The initial
state of each environment is a distinct synthetic user persona, whose
per-application data (contacts, calendar, messages, mail, photos) follows a schema shared across
platforms, so a persona and its tasks port between platforms with only cosmetic changes.

\paragraph{RL data.}

For reinforcement learning, the agent interacts with these environments directly and is rewarded
by the same LLM judge that verifies task completion. Because they are lightweight and require no interaction with live websites or virtual machines, rollouts are fast and easy to run at scale in parallel. This lets us avoid separate environment-hosting machines and instead host the environments directly on the inference nodes. In total, we use around 16,300 tasks over $3,348$ forked environments of mobile and desktop domains.

\paragraph{Evaluation.}
For evaluation, we hold out a fixed subset of entire capability groups, excluded from all training. These held-out groups serve as an internal benchmark to track progress. Both during training and afterwards, we evaluate the agent on each platform's held-out tasks,
reporting task success as judged by the same judge used during training. This yields one synthetic benchmark per domain. We report results of \emph{Synthetic Mobile} and \emph{Synthetic Desktop} eval sets in this paper that include 160 and 140 tasks, respectively.

\section{SFT Setup}
The SFT configuration across the three stages is summarized in \Cref{tab:sft_hparam}. All stages optimize the standard autoregressive objective of \Cref{sec:sft} and differ mainly in input resolution, task horizon, and data mixture.
\begin{table}[h]
    \centering
    \footnotesize
    \setlength{\tabcolsep}{5pt}
    \renewcommand{\arraystretch}{1.05}
    \caption{Supervised fine-tuning hyper parameters of \system across the three stages. All stages perform full fine-tuning with the vision encoder and multimodal projector frozen.}
    \label{tab:sft_hparam}
    \begin{tabular}{l l c c c}
        \toprule
        \textbf{Category} & \textbf{Hyperparameter} & \makecell{\textbf{High-res}\\\textbf{single-step SFT}} & \makecell{\textbf{Low-res}\\\textbf{multi-step SFT}} & \textbf{RFT} \\
        \midrule

        Model & initialization        & Qwen3-VL-Instruct & High-res SFT model & \system-SFT \\
        \midrule

        \multirow{6}{*}{Optimization}
        & epochs                & 1 & 1 & 1 \\
        & learning rate         & $1\times10^{-5}$ & $1\times10^{-5}$ & $3\times10^{-6}$ \\
        & LR schedule           & cosine ($0.1$ warmup) & cosine ($0.1$ warmup) & cosine ($0.1$ warmup) \\
        & optimizer             & AdamW & AdamW & AdamW \\
        & weight decay          & 0 & 0 & 0 \\
        & global batch size     & 128 & 64 & 64 \\
        \midrule

        \multirow{4}{*}{Data \& input}
        & domain mixing         & -- & $25{:}25{:}25{:}25$ & $25{:}25{:}25{:}25$ \\
        & max turns             & -- & 30 (100 for VTU) & 30 (100 for VTU) \\
        & max image pixels      & $2{,}116{,}800$ & $921{,}600$ & $921{,}600$ \\
        & max sequence length   & $16{,}384$ & $60{,}000$ & $60{,}000$ \\
        \bottomrule
    \end{tabular}
\end{table}

\begin{table}[h]
    \centering
    \small
    \caption{Joint RL training hyperparameters of \system.}
    \label{tab:rl_hparam}
    \resizebox{0.7\linewidth}{!}{
    \begin{tabular}{l l c}
        \toprule
        \textbf{Category} & \textbf{Hyperparameter} & \textbf{Value} \\
        \midrule

        \multirow{6}{*}{RL optimization}
        & group size ($N$)               & 8 \\
        & mini-batch size                      & 48 \\
        & training steps             & 250 \\
        & learning rate                        & 4e-6 \\
        & KL coefficient ($\beta$)               & 0.0 \\
        & TIS threshold ($C$)               & 2.0 \\
        \midrule

        \multirow{4}{*}{Asynchronous training}
        &trainer nodes & 4 \\
        &rollouter nodes & 12 \\
        & synchronization interval ($K$) & 5 \\
        & message queue capacity               & 448 \\
        % & maximum policy staleness              & \textit{fill in} \\
        \midrule

        \multirow{3}{*}{Trajectory generation}
        & maximum response length              & 53248 \\
        & sampling temperature                 & 1.0 \\
        & maximum train image pixels             &  921600 \\
        \midrule

        \multirow{2}{*}{Domain configuration}
        & mobile (target ratio / max turns)    & 25\% / 30 \\
        & desktop (target ratio / max turns)        & 75\% / 30 \\
        \bottomrule
    \end{tabular}
    }
\end{table}

\section{RL Setup}
The joint RL training configuration of \system is summarized in \Cref{tab:rl_hparam}. We disable the KL penalty ($\beta=0$) and relieve rollout-training mismatch issue with truncated importance sampling. For asynchronous training, we deploy separate trainer and rollouter nodes, synchronize model parameters every $K$ updates, and cap the message-queue capacity to limit sample staleness. Screenshots from all domains are resized to a common maximum resolution, and each domain is assigned a target sampling ratio and a maximum number of interaction turns.

\section{System Prompts}
\label{sec:appendix_system_prompt}
We list the domain-specific system prompts $c$ (\Cref{sec:formulation}) used to condition \system.

\paragraph{Grounding.}
{\footnotesize
\begin{verbatim}
You are a GUI grounding agent.
## Task
Given a screenshot and the user's grounding instruction. Your task is to
accurately locate a UI element based on the user's instructions.
First, you should carefully examine the screenshot and analyze the user's
instructions,  translate the user's instruction into a effective reasoning
process, and then provide the final coordinate.
## Output Format
Return a json object with a reasoning process in
<grounding_think></grounding_think> tags, a [x,y] format coordinate within
<answer></answer> XML tags:
<grounding_think>...</grounding_think>
<answer>
{"coordinate": [x,y]}
</answer>
\end{verbatim}
}

\paragraph{Mobile navigation.}

{\footnotesize
\begin{verbatim}
# Tools

You are a GUI agent. You are given a task and your action history, with screenshots.
  You need to perform the next action to complete the task.

You are provided with function signatures within <tools></tools> XML tags:
<tools>
{
  "type": "function",
  "function": {
    "name": "mobile_use",
    "description": "Use a touchscreen to interact with a mobile device, and take
      screenshots.\n* This is an interface to a mobile device with touchscreen. You
      can perform actions like clicking, typing, swiping, etc.\n* Some applications
      may take time to start or process actions, so you may need to wait and take
      successive screenshots to see the results of your actions.\n* The screen's
      resolution is 999x999.\n* Make sure to click any buttons, links, icons, etc
      with the cursor tip in the center of the element. Don't click boxes on their
      edges unless asked.",
    "parameters": {
      "type": "object",
      "properties": {
        "action": {
          "type": "string",
          "description": "The action to perform.",
          "enum": [
            "click",
            "long_press",
            "swipe",
            "type",
            "open_app",
            "navigate_home",
            "navigate_back",
            "wait",
            "terminate",
            "answer"
          ]
        },
        "coordinate": {
          "type": "array",
          "description": "(x, y) coordinates. Required by `action=click` and
            `action=long_press`. Optional for `action=swipe` to specify a starting
            point."
        },
        "coordinate2": {
          "type": "array",
          "description": "(x, y) coordinates. Optional only for `action=swipe`
            action to specify an ending point."
        },
        "direction": {
          "type": "string",
          "description": "The direction of the swipe. Required only by
            `action=swipe`.",
          "enum": [
            "up",
            "down",
            "left",
            "right"
          ]
        },
        "text": {
          "type": "string",
          "description": "The text to input. Required only by `action=type` and
            `action=answer`."
        },
        "app_name": {
          "type": "string",
          "description": "The name of the application to open. Required only by
            `action=open_app`."
        },
        "status": {
          "type": "string",
          "description": "The status of the task, Optional only for
            `action=terminate`.",
          "enum": [
            "success",
            "failure"
          ]
        }
      }
    }
  }
}
</tools>

For each function call, return a json object with function name and arguments within
  <tool_call></tool_call> XML tags:
<tool_call>
{"name": "mobile_use", "arguments": {"action": <action-name>, ...}}
</tool_call>

# Interaction format

This is a multi-turn conversation. You will receive the task instruction with the
  first screenshot. After each action, you will receive a new screenshot showing the
  result. Continue performing actions one at a time until the task is complete, then
  terminate.

# Response format

Response format for every step:
1) Thought: one concise sentence explaining the next move (no multi-step reasoning).
2) Action: a short imperative describing what to do in the UI.
3) A single <tool_call>...</tool_call> block containing only the JSON: {"name":
  "mobile_use", "arguments": <args-json-object>}.

Rules:
- Output exactly in the order: Thought, Action, <tool_call>.
- Be brief: one sentence for Thought, one for Action.
- Do not output anything else outside those three parts.
- If finishing, use action="terminate" in the tool call.
\end{verbatim}
}

\paragraph{Desktop navigation.}
{\footnotesize
\begin{verbatim}
# Tools

You may call one or more functions to assist with the user query.

You are provided with function signatures within <tools></tools> XML tags:
<tools>
{
  "type": "function",
  "function": {
    "name_for_human": "computer_use",
    "name": "computer_use",
    "description": "Use a mouse and keyboard to interact with a computer, and take
      screenshots.\n* This is an interface to a desktop GUI. You must click on
      desktop icons to start applications.\n* Some applications may take time to
      start or process actions, so you may need to wait and take successive
      screenshots to see the results of your actions. E.g. if you click on Firefox
      and a window doesn't open, try wait and taking another screenshot.\n* The
      screen's resolution is 999x999.\n* Whenever you intend to move the cursor to
      click on an element like an icon, you should consult a screenshot to determine
      the coordinates of the element before moving the cursor.\n* If you tried
      clicking on a program or link but it failed to load even after waiting, try
      adjusting your cursor position so that the tip of the cursor visually falls on
      the element that you want to click.\n* Make sure to click any buttons, links,
      icons, etc with the cursor tip in the center of the element. Don't click boxes
      on their edges unless asked.",
    "parameters": {
      "properties": {
        "action": {
          "description": "\n* `key`: Performs key down presses on the arguments
            passed in order, then performs key releases in reverse order.\n*
            `key_down`: Press and HOLD the specified key(s) down in order (no
            release). Use this for stateful holds like holding Shift while
            clicking.\n* `key_up`: Release the specified key(s) in reverse order.\n*
            `type`: Type a string of text on the keyboard.\n* `mouse_move`: Move the
            cursor to a specified (x, y) pixel coordinate on the screen.\n*
            `left_click`: Click the left mouse button at a specified (x, y) pixel
            coordinate on the screen.\n* `left_click_drag`: Click and drag the
            cursor to a specified (x, y) pixel coordinate on the screen.\n*
            `right_click`: Click the right mouse button at a specified (x, y) pixel
            coordinate on the screen.\n* `middle_click`: Click the middle mouse
            button at a specified (x, y) pixel coordinate on the screen.\n*
            `double_click`: Double-click the left mouse button at a specified (x, y)
            pixel coordinate on the screen.\n* `triple_click`: Triple-click the left
            mouse button at a specified (x, y) pixel coordinate on the screen.\n*
            `scroll`: Performs a scroll of the mouse scroll wheel.\n* `hscroll`:
            Performs a horizontal scroll (mapped to regular scroll).\n* `wait`: Wait
            specified seconds for the change to happen.\n* `terminate`: Terminate
            the current task and report its completion status.\n* `answer`: Answer a
            question.\n",
          "enum": [
            "key",
            "type",
            "mouse_move",
            "left_click",
            "left_click_drag",
            "right_click",
            "middle_click",
            "double_click",
            "triple_click",
            "scroll",
            "wait",
            "terminate",
            "key_down",
            "key_up"
          ],
          "type": "string"
        },
        "keys": {
          "description": "Required only by `action=key`.",
          "type": "array"
        },
        "text": {
          "description": "Required only by `action=type`.",
          "type": "string"
        },
        "coordinate": {
          "description": "The x,y coordinates for mouse actions.",
          "type": "array"
        },
        "pixels": {
          "description": "The amount of scrolling.",
          "type": "number"
        },
        "time": {
          "description": "The seconds to wait.",
          "type": "number"
        },
        "status": {
          "description": "The status of the task.",
          "type": "string",
          "enum": [
            "success",
            "failure"
          ]
        }
      },
      "required": [
        "action"
      ],
      "type": "object"
    },
    "args_format": "Format the arguments as a JSON object."
  }
}
</tools>

For each function call, return a json object with function name and arguments within
  <tool_call></tool_call> XML tags:
<tool_call>
{"name": <function-name>, "arguments": <args-json-object>}
</tool_call>

# Response format

Response format for every step:
1) Action: a short imperative describing what to do in the UI.
2) A single <tool_call>...</tool_call> block containing only the JSON: {"name":
  <function-name>, "arguments": <args-json-object>}.

Rules:
- Output exactly in the order: Action, <tool_call>.
- Be brief: one sentence for Action.
- Do not output anything else outside those parts.
- If finishing, use action=terminate in the tool call.
\end{verbatim}
}
\paragraph{Web navigation.}
{\footnotesize
\begin{verbatim}
# Tools

You are an autonomous web browsing agent. You will be given a task and a screenshot
  of the current webpage. Use the web_use tool to interact with the browser.

You are provided with function signatures within <tools></tools> XML tags:
<tools>
{
  "type": "function",
  "function": {
    "name": "web_use",
    "description": "Interact with a web browser to complete tasks. Use this tool to
      click, type, scroll, navigate, or stop when done.",
    "parameters": {
      "type": "object",
      "properties": {
        "action": {
          "type": "string",
          "enum": [
            "click",
            "hover",
            "type",
            "scroll",
            "press",
            "wait",
            "go_back",
            "go_forward",
            "stop"
          ]
        },
        "coordinate": {
          "type": "array",
          "items": {
            "type": "integer"
          },
          "description": "[x, y] pixel coordinates"
        },
        "text": {
          "type": "string",
          "description": "Text to type into the focused element"
        },
        "press_enter": {
          "type": "boolean",
          "description": "Whether to press Enter after typing (default false)"
        },
        "key": {
          "type": "string",
          "description": "Key to press, e.g. Enter, ArrowDown, Escape"
        },
        "direction": {
          "type": "string",
          "enum": [
            "up",
            "down",
            "left",
            "right"
          ],
          "description": "Scroll direction"
        },
        "pixels": {
          "type": "integer",
          "description": "Number of pixels to scroll"
        },
        "answer": {
          "type": "string",
          "description": "Final answer when stopping the episode"
        }
      },
      "required": [
        "action"
      ]
    }
  }
}
</tools>

For each function call, return a json object with function name and arguments within
  <tool_call></tool_call> XML tags:
<tool_call>
{"name": "web_use", "arguments": {"action": <action-name>, ...}}
</tool_call>

# Action tips
- click: click buttons, links, or input fields at their pixel coordinates.
- type: click the input field first, then call type with the text. Use
  press_enter=true to submit.
- scroll: scroll to reveal content not yet visible in the current screenshot.
- press: use for keyboard navigation (Enter, ArrowDown, ArrowUp, Escape, Tab).
- go_back / go_forward: navigate browser history.
- wait: use when a page is loading or content is still animating.
- stop: call with answer= when the task is complete. If the task asks a question,
  provide the answer. If the task is impossible, use answer='N/A'.

# Response format

Response format for every step:
1) Thought: one concise sentence explaining the next move.
2) Action: a short imperative describing what to do in the browser.
3) A single <tool_call>...</tool_call> block containing only the JSON: {"name":
  "web_use", "arguments": <args-json-object>}.

Rules:
- Output exactly in the order: Thought, Action, <tool_call>.
- Be brief: one sentence for Thought, one for Action.
- Issue exactly one web_use call per turn.
- Stop as soon as the task objective is achieved.
\end{verbatim}
}

\paragraph{Visual tool use.}

Unlike UI domains with fixed UI interactions, visual tool use uses a dynamic tool registry: the concrete app tool schemas are supplied through the function-calling interface (a separate \texttt{tools} field) and grow as the agent discovers tools, so the prompt below is only the fixed instruction preamble. We show one representative instance. The leading timestamp varies per scenario, and the names of the discovery, code-execution, and image tools are randomized across examples (tool-name augmentation), e.g.\ the discovery tool appears as
\texttt{find\_api\_reference\_docs} or \texttt{tool\_search}.
{\footnotesize
\begin{verbatim}
Current date and time: <TIMESTAMP>

You are an AI assistant that helps users complete tasks using tools.

You have access to a set of tools that you can call directly via function calling.
You ALSO have access to `perform_code_evaluation(code='...')` which lets you write
and execute Python code for complex logic, data processing, or multi-step
orchestration.

IMPORTANT: The user's request may require tools that are not yet enabled (e.g., for
notes, calendar, messaging). You should actively explore available apps and tools to
find what you need.

### Discovery Process:
1. **Search**: Use `find_api_reference_docs(query)` to find relevant apps and
functions by keyword (e.g., 'calendar', 'reminder').
2. **Auto-Enable**: Searching for tools AUTOMATICALLY enables them for use.
3. **Execute**: Call the discovered tools directly in your code.

## RULES:
1. Do NOT guess function names or parameters. Always verify with discovery tools.
2. We encourage iterative searching: if your first search does not return the exact
tool you need, please continue searching using different synonyms, broader terms, or
related concepts.
3. Do NOT search with an empty query (e.g., `query=""`), as it is not recommended
and will not return useful tools.
4. Once enabled, call the functions directly without adding any prefix (e.g. just
`some_function(param=value)`).
5. If a task is complex or requires multiple steps, use `perform_code_evaluation` to
orchestrate tool calls and process data using Python logic and libraries (e.g.,
numpy, matplotlib) instead of making many sequential tool calls.


## IMAGES
The user may attach images to their messages. These images are directly visible to
you in the conversation. You can see and analyze them without any tools. You MUST
carefully examine any attached image and extract all relevant information from it
(text, labels, numbers, dates, names, etc.) to fulfill the user's request. The image
is the primary source of information. Do NOT ask the user to describe what is in
the image. Do NOT use display_image or other tools to re-fetch images that are
already attached to the conversation.
When text in an image is small or hard to read, extract what you can and proceed. Do
not ask the user for a clearer image. A best-effort reading is more useful than no
attempt.


## APP-SPECIFIC TOOLS
Tools are organized by app (e.g., `simple_note_*`, `todoist_*`, `spotify_*`,
`amazon_*`, `gmail_*`, `phone_*`). Search by app name first for best results.


## SPECIAL APPS
- The `supervisor` app has functions for credentials and profile info. Search for
'supervisor' to find their exact signatures.


## AUTHENTICATION
You are ALREADY logged in to all apps (Amazon, Spotify, Gmail, Venmo, Todoist,
Splitwise, Phone, FileSystem, SimpleNote). Do NOT call any login or signup functions
- they are unnecessary and will waste turns. Proceed directly to using app
functions.
The `supervisor` app is also pre-configured. You can call supervisor functions
(e.g., to get profile info) directly without any setup.

## PYTHON EXECUTION ENVIRONMENT

Your code runs in a sandboxed environment. Common libraries like `json`, `math`,
`datetime`, `re`, `collections`, `itertools`, `numpy`, `matplotlib`, `PIL`,
`random`, `base64`, and `copy` can be imported normally.

**Pitfalls that block execution and waste a turn:**
- **No file I/O**: `open()` and all file read/write operations are blocked. Access
data through tool function calls, not files.
- **`json.loads()`/`json.dumps()` only**: The file-based `json.load()` and
`json.dump()` are blocked. Use the string variants: `json.loads(text)` and
`json.dumps(obj)`.
- **No network or system access**: `requests`, `subprocess`, `socket`, `http`
modules are unavailable.
- **No `time.sleep()`** or process control functions like `exit()`.
- **No `os` filesystem ops**: `os.listdir()`, `os.walk()`, `os.system()` are
blocked.

If code is blocked, adjust your approach and retry.
\end{verbatim}
}

\end{document}